\documentclass{article}
\usepackage{subcaption}

\usepackage{styles/user_commands}     
\usepackage{standalone}  
\usepackage{multirow,graphicx, xcolor}
\usepackage{booktabs}
\usepackage{url}

\usepackage{makecell}
\usepackage{tikz}
\usetikzlibrary{positioning,chains, matrix}
\usepackage[utf8]{inputenc} 
\usepackage[T1]{fontenc}    
\usepackage{url}          
\usepackage{booktabs}      
\usepackage{amsfonts}     
\usepackage{nicefrac}      
\usepackage{microtype}      
\usepackage{xcolor}   
\usepackage{subfiles}
\usepackage{dsfont}  
\usepackage{mathtools}
\usepackage{graphicx}
\usepackage{hyperref}
\usepackage{changepage}
\usepackage{mdframed}

\usepackage[accepted] {icml2024}
\usepackage[round,authoryear]{natbib}
\icmltitlerunning{Deep Functional Factor Models}
\newtheorem{theorem}{Theorem}

\begin{document}
%\title{%\bf \Large 
%DF\textsuperscript{2}M: An Explainable Deep Bayesian Nonparametric Model for High-Dimensional Functional Time Series
%}

%\author{Yirui Liu$^{1}$ \quad Xinghao Qiao$^{1}$  \quad Yulong Pei$^{2}$  \quad Liying Wang$^{3}$  \\$^{1}$London School of Economics and Political Science \\ $^{2}$Eindhoven University of Technology   \quad  $^{3}$University of Liverpool} 

%\maketitle

\twocolumn[

\icmltitle
{Deep Functional Factor Models: Forecasting High-Dimensional Functional Time Series via Bayesian Nonparametric Factorization
}
%{DF\textsuperscript{2}M: An Explainable Deep Bayesian Nonparametric Model for High-Dimensional Functional Time Series}
%\icmlsetsymbol{equal}{*}

\begin{icmlauthorlist}
\icmlauthor{Yirui Liu}{jpm,lse}
\icmlauthor{Xinghao Qiao}{hku}
\icmlauthor{Yulong Pei}{jpm}
\icmlauthor{Liying Wang}{liv}
\end{icmlauthorlist}

\icmlaffiliation{jpm}{%Chief Data \& Analytics Office, 
J.P. Morgan}
\icmlaffiliation{lse}{London School of Economics and Political Science}
\icmlaffiliation{hku}{Faculty of Business and Economics, The University of Hong Kong}
%\icmlaffiliation{sch}{Eindhoven University of Technology}
\icmlaffiliation{liv}{Management School, University of Liverpool}

\icmlcorrespondingauthor{Xinghao Qiao}{xinghaoq@hku.hk}
\icmlkeywords{Bayesian nonparametric, Deep learning, Factor model, Functional time series, Gaussian process, Indian buffet process, Sparse variational inference}

\vskip 0.3in
]

\printAffiliationsAndNotice{}

\begin{abstract}
This paper introduces the Deep Functional Factor Model (DF\textsuperscript{2}M), a Bayesian nonparametric model designed for analysis of high-dimensional functional time series. DF\textsuperscript{2}M is built upon the Indian Buffet Process and the multi-task Gaussian Process, incorporating a deep kernel function that captures non-Markovian and nonlinear temporal dynamics. Unlike many black-box deep learning models, DF\textsuperscript{2}M offers an explainable approach to utilizing neural networks by constructing a factor model and integrating deep neural networks within the kernel function. Additionally, we develop a computationally efficient variational inference algorithm to infer DF\textsuperscript{2}M. Empirical results from four real-world datasets demonstrate that DF\textsuperscript{2}M provides better explainability and superior predictive accuracy compared to conventional deep learning models for high-dimensional functional time series.
\end{abstract}

%\noindent {\small{\it Some keywords}: Bayesian nonparametric, Deep learning, Factor model, Functional time series, Gaussian process, Indian buffet process, Sparse variational inference }

\section{Introduction}
\label{sec:intro}
Functional time series refers to a sequential collection of functional objects that exhibit temporal dependence, and this area of study has garnered increasing attention in recent years. With the advancements in data collection technology and computational power, high-dimensional datasets containing numerous functional time series have become more prevalent. Examples of such data include annual age-specific mortality rates for different countries, daily energy consumption curves from various households, and cumulative intraday return trajectories for hundreds of stocks. These datasets can be represented as $p$-dimensional functional time series $\bY_t (\cdot) = \big(Y_{t1}(\cdot), \dots, Y_{tp}(\cdot) \big)^{\T}$, where each $Y_{tj}(\cdot)$ is a random function defined on a compact interval $\cU$. The number of functional variables $p$ is comparable to, or even larger than, the number of temporally dependent observations $n$. Analyzing high-dimensional functional time series presents a challenging task, as it necessitates the use of dimension reduction techniques to address the high-dimensional problem, functional approaches to handle the infinite-dimensional nature of curve data, and time series modeling methods to capture the temporal dependence.

Several statistical methods have been proposed to address these challenges, such as those presented in \cite{gao2019, chen2022, fang2022, chang2023, guoqiao2023, zhou2023}. However, these approaches often assume the existence of linear and Markovian dynamics over time, which may fail to accurately capture the complex nonlinear or non-Markovian temporal dependence that often arises in real-world scenarios.

On the other hand, while deep learning has achieved impressive results in computer vision and natural language processing (NLP) \cite[]{guo2016deep, he2016, vaswani_attention_2017, torfi2020}, applying deep neural networks directly to handle high-dimensional functional time series is challenging. One major issue when dealing with time series data is that deep learning is a general black-box method that lacks explainability, thus making it difficult to understand the cross-sectionally and serially correlated relationships. However, explainability is crucial in many applications. For instance, in finance, healthcare, and climate change, the accuracy and reliability of a model's predictions have significant impacts on business decisions, patient outcomes, or environmental safety, respectively. Additionally, the non-stationarity of data and the large number of parameters in deep neural networks pose extra challenges during training.

In this paper, we propose an explainable approach called the deep functional factor model (DF\textsuperscript{2}M), which has the ability to discover nonlinear and non-Markovian dynamics in high-dimensional functional time series. Developed as a Bayesian nonparametric model, DF\textsuperscript{2}M employs a functional version of a factor model for dimension reduction, incorporates an Indian buffet process prior in the infinite-dimensional loading matrix to encourage column sparsity \cite[]{guo2021a}, utilizes a functional version of a Gaussian process dynamical model to capture temporal dependence within latent functional factors, and employs deep neural networks to construct the temporal kernel.

DF\textsuperscript{2}M offers several advantages for the analysis of high-dimensional functional time series. (i) Firstly, it facilitates a more intuitive understanding of the underlying data structure by representing observed curve variables using a smaller set of latent functional factors. This enhances model explainability and provides a clear and interpretable mapping of relationships between variables, which is
crucial for decision-making and subsequent analysis. (ii) Secondly, DF\textsuperscript{2}M is capable of discovering non-Markovian and nonlinear temporal dependence in the functional latent factor space. This enables more accurate predictions of future values.
(iii) Lastly, DF\textsuperscript{2}M offers a flexible framework that combines modern sequential deep neural networks with a backbone Bayesian model. This allows for the utilization of sequential deep learning techniques such as gated recurrent unit (GRU) \cite[]{cho2014properties}, long short-term memory (LSTM) \cite[]{hochreiter1997long}, and attention mechanisms \cite[]{vaswani_attention_2017}. 

\section{Preliminaries}
\subsection{Indian Buffet Process}
\label{subsec:indian_buffet_process}
The Indian buffet process (IBP) \cite[]{griffiths2011} is a probability distribution over a sparse binary matrix with a finite number of rows and an infinite number of columns. The matrix $\bZ$, generated from the IBP with parameter $\alpha$, is denoted as $\bZ \sim \text{IBP}(\alpha)$, where $\alpha$ controls the column sparsity of $\bZ$. 

IBP can be explained using a   metaphor that customers sequentially visit a buffet and choose dishes. The first customer samples a number of dishes based on Poisson$(\alpha)$. Subsequent the $i$-th consumer, in turn, samples each previously selected dish with a probability proportional to its popularity ($m_k/i$ for dish $k$), and also tries new dishes following Poisson$(\alpha/i)$.

It is worth noting that the distribution remains exchangeable with respect to the customers, meaning that the distribution is invariant to the permutations of the customers. The Indian buffet process admits a stick-breaking representation
as $v_j \mid \alpha \sim \text{Beta} (\alpha, 1)$ independently for $j=1,2,\dots$, $\ w_k = \prod_{j=1}^kv_j$ for $k=1,2,\dots$, and $Z_{ik} \mid w_k \sim \text{Bernoulli} (w_k)$ independently for $i=1,\dots, n$, and the IBP is then defined as $\bZ={(Z_{ik})}_{1\leq i \leq n, k \geq 1}$. The stick-breaking representation is frequently used in the inference for IBP.

\subsection{Gaussian Process}
A Gaussian process $X(\cdot)$, defined on a compact interval $\cU$, is a continuous stochastic process characterized by the fact that every finite collection of its values, $X(u_{1}),\dots,X(u_{L})$ with $u_{1},\dots,u_{L} \in \cU$, belongs to an $L$-dimensional multivariate Gaussian distribution \cite[]{williams2006}. This means that a Gaussian process is completely determined by its mean function $m(u) = \E\big[X(u)\big]$ and its covariance function $\kappa (u, v) = \text{Cov}\big(X (u) , X (v)\big) = \E \Big[\big(X(u) - m(u)\big) \big(X(v) -m (v)\big)\Big]$ for any $u, v \in \cU$.

The covariance function, also known as the kernel function in machine learning literature, specifies the correlation between values at distinct points. Examples include the squared exponential kernel $\kappa(u, v) = \exp \left( - |u-v |^2/\ell^2 \right)$ and the Ornstein--Uhlenbeck kernel $\kappa(u,v)=\exp \left(- |u-v  |/\ell  \right)$, where $\ell$ is the length-scale parameter. Additionally, the kernel function can be made more complex using the kernel trick \cite[]{hofmann2008} by rewriting it as $\kappa(u,v) =\big\langle \phi(u), \phi(v) \big\rangle$, where $\langle \cdot, \cdot \rangle$ denotes the inner product, and the feature function $\phi(\cdot)$ maps $x$ into a feature space. As $\phi(\cdot)$ can be an arbitrary function (linear or nonlinear), the Gaussian process offers considerable flexibility in modeling complex patterns in the data. 

Furthermore, a multi-task Gaussian process (MTGP) \cite[]{bonilla2007} can be employed to model vector-valued random fields. It is defined as $\mathbf{X}(\cdot) = \big(X_1(\cdot), \dots, X_M(\cdot)\big)^{\T}$, where $X_1(\cdot),\dots, X_M(\cdot)$ are $M$ Gaussian processes defined on $\mathcal{U}$. The covariance function between the $l$-th and $k$-th task is given by $\text{Cov}\big(X_l(u), X_k(v)\big) = \Sigma_{lk}\kappa(u, v)$, where $\mathbf{\Sigma} = \{ \Sigma_{lk}\}_{1 \leq l, k \leq M}$ is a positive semi-definite matrix encoding the similarities between pairs of tasks. The MTGP can effectively capture inter-task correlations and improve predictions \cite[]{moreno2018}.

\begin{figure*}[t] % \begin{figure*}[t]
\centering
\includegraphics[width=0.68\linewidth]{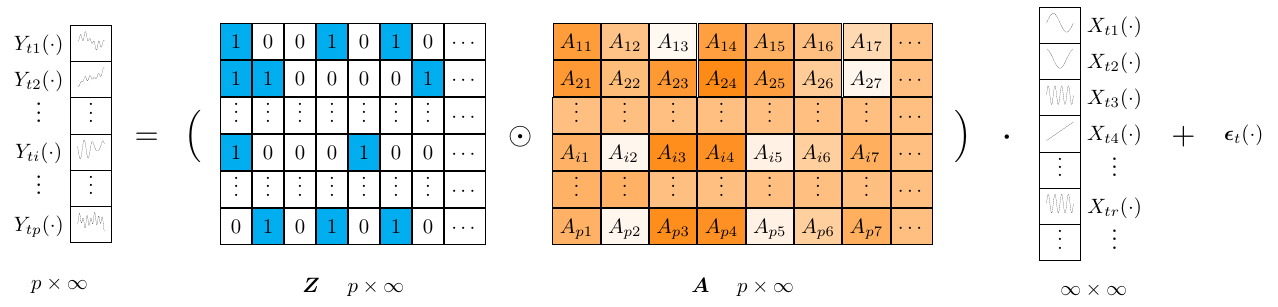}
% -5pt
\caption{Sparse functional factor model. The blue (or white) cells in $\boldsymbol{Z}$ indicate $1$ (or $0$), while the darker (or lighter) shades of orange in $\boldsymbol{A}$ represent larger (or smaller) values.}
\label{fig:the_model_plot}
% -10pt
\end{figure*}
\subsection{Sequential Deep Learning}
Deep learning methods, widely used in computer vision, NLP, and reinforcement learning, have become increasingly popular for time series prediction as well \cite[]{lim2021}. In particular, recurrent neural networks (RNN) and attention mechanisms, commonly used for sequence prediction tasks in NLP, can be adapted for temporal forecasting tasks in time series data. A multivariate time series can be modeled recursively in RNN as $\mathbf{x}_t = g_{\text{dec}} (\mathbf{h}_t)$ and $\mathbf{h}_t = g_{\text{enc}} (\mathbf{h}_{t-1}, \mathbf{x}_{t-1})$, where $\mathbf{h}_t$ is a latent variable and $g_{\text{dec}}$ and $g_{\text{enc}}$ are the decoder and encoder functions, respectively. Two renowned RNN models, LSTM and GRU, are designed to learn long-range dependencies in a sequence. For simplicity, we denote their encoder functions as $\mathbf{h}_{t} = \text{LSTM}(\mathbf{x}_{1:t})$ and $\mathbf{h}_{t} = \text{GRU}(\mathbf{x}_{1:t})$, respectively, where $\mathbf{x}_{1:t} = (\mathbf{x}_{1}, \dots, \mathbf{x}_{t})$. Moreover, attention mechanisms, which have achieved state-of-the-art performance in NLP tasks, can also be utilized to model time series data. Unlike RNNs, attention mechanisms directly aggregate information from multiple time steps in the past. Attention mechanisms can be expressed as $\mathbf{h}_t = \sum_{i=1}^{t-1} {\omega(\mathbf{k}_t, \mathbf{q}_{\tau}) \mathbf{v}_{t-\tau}}$, where key $\mathbf{k}_t$, query $\mathbf{q}_t$, and value $\mathbf{v}_t$ are intermediate representations generated by linear or nonlinear transformations of $\mathbf{x}_t$. We denote such attention mechanisms as $\mathbf{h}_{t} = \text{ATTN}(\mathbf{x}_{1:t})$. See detailed structures for both RNNs and attention mechanisms in Appendix~\ref{app:dl_modules}.

\section{Deep Functional Factor model}

\subsection{Sparse Functional Factor Model}
First, we propose a functional factor model from the Bayesian perspective,
\begin{equation}
\label{eq:the_model}
    \bY_t(\cdot) = (\bZ \odot \bA) \bX_t(\cdot) + \bepsilon_t(\cdot), \ \ t=1,\dots,n.
\end{equation}
The observed functional time series is denoted as $\bY_t(\cdot)$, and the binary matrix $\bZ$ is sampled from the Indian buffet process, $\bZ \sim \text{IBP}(\alpha)$. The Hadamard (elementwise) product is represented by $\odot$. The loading weight matrix $\bA$ has elements $A_{tr} \sim \text{Normal}(0, \sigma_{A}^2)$ for any $r \in \N^+$ independently. The latent functional factor time series are denoted as $\bX_t(\cdot)=\big(X_{t1}(\cdot), X_{t2}(\cdot), \dots, X_{tr}(\cdot), \dots \big)^{\T}$, and the idiosyncratic component is denoted as $\bepsilon_t(\cdot)$, which follows a Gaussian distributed white noise process on a scale $\sigma_\epsilon$.

In this framework, we do not specify the number of latent factors. Instead, $\bY_t(\cdot)$, $\bZ$, and $\bA$ can be regarded as $p \times \infty$ matrices, and $\bX_t(\cdot)$ as an infinite-dimensional vector of functions, or heuristically, a $\infty \times \infty$ matrix. The dimension reduction framework in equation~\eqref{eq:the_model} is illustrated in Figure~\ref{fig:the_model_plot}. This Bayesian nonparametric factor model allows for a potentially unlimited number of latent factors, eliminating the need to specify a fixed dimensionality of the factor space. The nonparametric approach introduces flexibility and provides a foundation for inferring the number of factors in the posterior distribution using nonparametric inference frameworks such as Gibbs sampling \cite[]{teh2006}, merge-split algorithm \cite[]{hughes_memoized_2013}, and conditional and adaptively truncated variational inference \cite[]{liu2020, eegnn2023}.

Additionally, the Indian buffet process can provide column sparsity \cite[]{vu2013} to $\bZ$ and, therefore, to the loading matrix $\bZ \odot \bA$. This implies that most elements in each row are zeros, as $w_k$ in the stick-breaking representation of the IBP approaches zero as $k$ increases. In the factor model, this column sparsity implies that each factor affects only a small fraction of functional variables \cite[]{guo2021a}, or equivalently, the factors are related to each other through a hierarchy \cite[]{griffiths2011}.

\subsection{Functional Gaussian Process Dynamical Model}
\label{subsec:gpdmfd}
Secondly, by projecting high-dimensional observations $\bY_t(\cdot)$ onto low-dimensional latent functional factors $\bX_t(\cdot)$, we can capture the sequential structure of the time series model through the factors. To model the temporal dependence of $\bX_t(\cdot)$, we adopt a Gaussian process over time to encode historical information. In particular, we design the covariance across factors $r$ and $l$ as follows. Let $\cX_t$ represent the historical information up to time $t$, and let $\cX$ be the space containing $\{\cX_t\}_{t \in \N}$. For any $u, v \in \cU$,
\begin{equation}
\label{eq:variance_product_model}
\text{Cov}\big(X_{tr}(u), X_{sl}(v)\big) = \kappa_{\cX}(\cX_{t-1}, \cX_{s-1}) \kappa_{\cU}(u, v)\mathbb{I}(r=l),
\end{equation}
where $t$ and $s$ indicate two time stamps, and $\kappa_{\cX}$ and $\kappa_{\cU}$ are the kernels defined on $\cX$ and $\cU$, respectively. The indicator function $\mathbb{I}(r=l)$ equals 1 if $r=l$ and 0 otherwise. The kernel $\kappa_{\cX}$ captures historical information from different periods and can be regarded as a temporal kernel. Similarly, the kernel $\kappa_{\cU}$ can be seen as a spatial kernel.

Therefore, $\bX_r(\cdot) = \big(X_{1r}(\cdot), \dots, X_{tr}(\cdot), \dots, X_{nr}(\cdot)\big)$ belongs to a multi-task Gaussian process \cite[]{bonilla2007} such that for any $u_1, \dots, u_L \in \cU$, $\text{vec} \big(\bX_r(u_1, \dots, u_L)\big) \sim \text{Normal}(\boldsymbol{0}, \bSigma_{\cX} \otimes \bSigma_{\cU}^u)$, where $\otimes$ denotes the Kronecker product, or equivalently $\bX_r(u_1, \dots, u_L)$ follows a matrix normal distribution \cite[]{dawid1981} with mean $\boldsymbol{0}$, row covariance matrix $\bSigma_{\cU}^u$, and column covariance matrix $\bSigma_{\cX}$.
Here, 
$
\bX_r(u_1, \dots, u_L) = 
\big[X_{tr}(u_j)\big]_{1 \leq t \leq n, 1 \leq j \leq L},
$
$
\bSigma_{\cX} = \big[\kappa_{\mathcal{X}}(\mathcal{X}_t, \mathcal{X}_s)\big]_{0 \leq t, s \leq n-1},
$ 
and 
$
\bSigma_{\cU}^u = 
\big[\kappa_{\mathcal{U}}(u_i, u_j)\big]_{1 \leq i, j \leq L}.
$

In Appendix~\ref{app:matrix_mormal}, we demonstrate the detailed relationship between the multi-task Gaussian process and the matrix normal distribution.
In the literature, the $n$-task Gaussian process is often used to refer to multiple outputs generated by the model, each corresponding to a specific timestamp in our time series setting.
For ease of expression, we denote the presented multi-task Gaussian process as
\begin{equation}
\label{eq:eq_model}
    \bX_r(\cdot) \sim  \text{MTGP}\big(\boldsymbol{0}, \kappa_{\cU}(\cdot, \cdot),  \kappa_{\cX}(\cdot,\cdot)\big).
\end{equation}
Importantly, the marginal distributions of latent factors, denoted as $\bX_1, \dots, \bX_r, \dots$, exhibit cross-sectional dependence due to their common temporal kernel $\kappa_{\cX}$, which incorporates historical information up to period $t-1$. This shared kernel promotes similarity and dependence across different time periods in the multi-task Gaussian process. However, when conditioned on $\cX_{t-1}$ and the corresponding kernel, $\bX_1, \dots, \bX_r, \dots$, become conditionally independent and Gaussian distributed. This conditionality arises from our approach, where the predictive temporal kernel exclusively relies on past information $\cX_{t-1}$ rather than current data $\cX_{t}$. Consequently, this approach enables forward-looking predictions based solely on historical data.

The proposed model can be seen as a functional variant of the Gaussian process dynamical model \cite[]{wang2005}. The connections between the two models can be found in Appendix A. Using the proposed model to capture the temporal dependence of functional time series has several advantages. First, the temporal kernel and spatial kernel can be separated, which allows for a closed form and computational convenience. Additionally, since the temporal kernel considers the entire historical information rather than just the latest state, the model can be non-Markovian. For example, by defining the temporal kernel as
$
\kappa(\mathcal{X}_{t-1}, \mathcal{X}_{s-1}) = \alpha_1 \int \mathbf{X}_{t-1}(u)^T \mathbf{X}_{s-1}(u) du + \alpha_2 \int \mathbf{X}_{t-2}(u)^T\mathbf{X}_{s-2}(u) du,
$ 
features from the last two periods can be incorporated.
Furthermore, nonlinearity can be introduced using the kernel trick by setting a nonlinear kernel function. This opens the possibility of constructing deep kernels, which will be discussed in Section~\ref{sec:deep_kernel}.

\subsection{Deep Temporal Kernels}
\label{sec:deep_kernel}
To capture the complex latent temporal structure, we use neural networks to construct the kernel function. However, compared to standard deep kernels  \cite[]{wilson2016, al2017, xue2019, li2019, watson2021,fortuin2022}, two extra steps are needed when applying deep kernels to functional time series.

Firstly, since $\mathbf{X}_t(\cdot)$ is a continuous process on $\mathcal{U}$, a mapping function $F: \mathcal{F} \rightarrow \mathbb{R}^d$ is required to map the infinite-dimensional Gaussian processes to $d$-dimensional vectors. Here, $\mathcal{F}$ represents the space of continuous functions defined on $\mathcal{U}$. Various approaches can be used for this mapping function, including pre-specified basis expansion, data-dependent basis expansion (such as functional principal component analysis and its dynamic variants  \cite[]{bathia2010, hormann2015}), adaptive functional neural network 
\cite[]{yao2021a}, or even a simple specification such as  $\bX_t(u_0, \dots, u_L)$ with $u_0, \dots, u_L \in \cU$.

Secondly, the $d$-dimensional vectors are used as inputs for deep neural networks, and the outputs generated by these networks are employed to construct kernel functions. Specifically, the input vector is transformed as
%% -13pt
\begin{equation}
\bh_t= H\big(F(\bX_{t-1}), F(\bX_{t-2}), \dots\big),
\end{equation}
where $\mathbf{X}_{t-1} = \left( X_{t-1, 1}, \dots, X_{t-1, r}, \dots\right)^T$, $F$ is the mapping function, and $H$ represents a sequential deep learning framework. Various deep neural network architectures can be utilized for this purpose, such as LSTM, GRU, and attention mechanisms, which have shown their effectiveness in modeling complex patterns and dependencies. Since the inputs for the temporal kernels are ordered sequences from $\mathcal{X}_{0}$ to $\mathcal{X}_{n-1}$, unidirectional deep neural networks should be used instead of bidirectional networks. The transformed representations $\mathbf{h}_t$ and $\mathbf{h}_s$ are then used to construct a kernel 
\begin{equation}
\kappa_{\cX}(\cX_{t-1}, \cX_{s-1}) = \kappa(\bh_t, \bh_s),
\end{equation}
where $\kappa(\cdot, \cdot)$ is a suitable kernel function, such as the squared exponential kernel or the Ornstein-Uhlenbeck kernel. It should be noted that the temporal kernel is related to the historical values of all the relevant factors and is shared across factors. This approach builds a time-dependent variant of a deep Gaussian process \cite[]{damianou2013}, where the kernel incorporates historical information.

To address overfitting, spectrum normalization can be applied, which effectively enforces a Lipschitz condition on the neural networks as suggested by \cite{miyato2018}. Notably, this Lipschitz condition on the model's output concerning its input ensures that similar inputs within the Gaussian process exhibit comparable distances in the kernel space. This implies that the inputs for $\kappa$ can reflect the distance between $\mathcal{X}_{t-1}$ and $\mathcal{X}_{s-1}$.

\subsection{The Imperative of Element Integration}
In summary, by combining the functional version of the sparse factor model, sequential deep learning kernel, and Gaussian process dynamical model, we define a probabilistic generative model for high-dimensional functional time series named the deep functional factor model (DF$^2$M).

Factorization is essential in this context. Directly feeding the original high-dimensional input into the kernel function is not viable. The issue arises because without the factorization of high-dimensional functional data, the kernel would have to handle extremely high-dimensional inputs using an excessive number of parameters, while in training time steps are limited. This situation is prone to overfitting, making it challenging to achieve accurate estimation. Additionally, the interpretability of our proposed method relies on these factors.

Applying a deep kernel  without utilizing an IBP with a finite number of latent factors poses difficulties. The challenge lies in determining the optimal number of latent factors. If there are too many factors or too few, it can affect the kernel distance. An excess of factors, often redundant and similar, can distort distance measurements, while an inadequate number of factors may miss critical distances, leading to inaccurate model representations.

The sequential deep learning kernel is essential for introducing non-Markovian and non-linear patterns. When combined with IBP and factorization, it effectively models temporal similarities among observations, enabling capture of high-dimensional functional data dynamics.

\section{Bayesian Inference for DF\textsuperscript{2}M}
\label{sec:inference}
\subsection{Sparse Variational Inference}
We adopt the variational inference framework to infer the proposed DF\textsuperscript{2}M. This algorithm approximates the posterior probability by maximizing the evidence lower bound (ELBO), which is equivalent to minimizing the Kullback--Leibler (KL) divergence between a variational distribution and true posterior distribution \cite[]{blei2017}. For DF\textsuperscript{2}M, with mean-field factorization assuming independence among the variational distributions for latent variables, its ELBO can be expressed as
%{-5pt}
\begin{equation}
\label{eq:elbo}
\begin{gathered}
\text{ELBO} = \E_q \Big[\log p(\bZ\mid \alpha) p(\bA \mid \sigma_A)  \\\prod_{t=1}^{n}p\big(\bY_t(\cdot) \mid \bX_t(\cdot), \bZ, \bA\big)\prod_{r\geq1} p\big(\bX_r(\cdot) \mid 
 \kappa_{\cX}, \kappa_{\cU} \big) \Big] \\ 
 %+ \log p(\kappa_{\cX})+ \log p(\kappa_{\cU})
 -   \E_q \Big[\log q(\bZ) q(\bA) \prod_{r\geq1} q\big(\bX_r(\cdot)\big)\Big].
\end{gathered}
\end{equation}
% -5pt
Using the stick-breaking representation of the Indian buffet process as in Section~\ref{subsec:indian_buffet_process}, we factorize the variational distribution for $\bZ$ as $q(v_j)=\text{Beta}(v_j;\tau_{j}^1, \tau_{j}^0)$ and $q(Z_{tj}) = \text{Bernoulli}(Z_{tj}; m_{tj})$. The corresponding variational distribution for $\bA$ is factorized as  $q(A_{tj}) = \text{Normal}(A_{tj}; \eta_{tj}, {\sigma}^2_{A, tj})$. 

To avoid singular matrix inversions and improve computational efficiency,  we propose a sparse variational inference approach for DF\textsuperscript{2}M based on \cite{titsias2009}.  Our method  introduces a set of inducing variables representing the values of the latent function at a small subset of points in $\cU$. Moreover,
%It has been demonstrated that the variational distribution maximizing ELBO of the Gaussian process is equivalent to the prior distribution conditional on the variational distribution of inducing variables.
%When performing inference on a multi-task Gaussian process, 
we adopt the approach of having common locations for the inducing variables across functional factors, as suggested by \cite{hamelijnck2021}. In other words, we utilize the same set of inducing points for all tasks, which can lead to further improvement in computational efficiency.  Consequently, the variational distribution for multi-task Gaussian process with inducing variables is defined as,
%{-5pt}
\begin{equation}
\label{eq:variatonal_conditional}
\begin{gathered}
    q\big(\bX_{r}(\cdot)\big)=p\Big(X_{1r}(\cdot), \dots, X_{nr}(\cdot) \mid X_{1r}(\bv), \dots, \\X_{nr}(\bv)
    , \kappa_{\cX}, \kappa_{\cU}\Big)\prod_{t=1}^{n}q\big(X_{tr}(\bv)\big),
\end{gathered}
\end{equation}
% -5pt
where $\bv=(v_1, \dots, v_K)^{\T}$ with $v_1, \dots, v_K \in \cU$ with $K$ being the number of inducing points. The variational distribution for the inducing variables is constructed as $q(X_{tr}(\bv)) = \text{Normal}(\bmu_{tr}, \bS_{tr})$. It is important to note that the conditional prior distribution for $X_r(\cdot)$, , which is the first term on the right-hand side of equation~\eqref{eq:variatonal_conditional}, cannot be factorized as $\prod_{t=1}^{n}p\big(X_{tr}(\cdot) \mid X_{tr}(\bv)\big)$ due to their temporal dependence. However,  by exploiting the setting of equation~\eqref{eq:variatonal_conditional}, the conditional prior distribution appears in both the variational and prior distributions and therefore can be cancelled. In Appendix~\ref{app:sparse_variational}, we derive that the ELBO in equation~\eqref{eq:elbo} can be simplified as
\begin{equation}
\begin{gathered}
\label{eq:elbo_rewrite}
%\begin{split}
\text{ELBO} = 
\sum_{t=1}^{n}\E_q \Big[ \log p\big(\bY_t(\cdot) \mid \bX_t(\cdot), \bZ, \bA\big) \Big] 
%+ \log p(\kappa_{\cX})+ \log p(\kappa_{\cU})
 \\-\text{KL}\big[q(\bZ) \parallel p(\bZ\mid \alpha)\big]    - \text{KL} \big[q(\bA)\parallel p(\bA \mid \sigma_A)
 \big]  \\- \sum_{r\geq1}  \text{KL} \Big[q\big(\bX_r(\bv)\big)\parallel p\big(\bX_r(\bv) \mid \kappa_{\cX}, \kappa_{\cU}\big)
 \Big],
%\end{split}
\end{gathered}
\end{equation}
% -5pt
where $\bX_r(\bv) = \big(X_{1r}(\bv) , \dots,X_{nr}(\bv)\big)$ with $X_{tr}(\bv)=\big(X_{tr}(v_1), \dots, X_{tr}(v_K)\big)^{\T}$ for $t=1,\dots, n$.  Furthermore, using the formula of the KL divergence between two multivariate Gaussian distributions,  we derive a closed form of the last term as
\begin{equation}
\begin{gathered}
\label{eq:divergence}
2\text{KL} \Big[q\big(\bX_r(\bv)\big)\parallel p\big(\bX_r(\bv) \mid \kappa_{\cX}, \kappa_{\cU}\big)  \Big]\\=\text{trace}\Big((\bSigma_{\cX}^{-1} \otimes {\bSigma_{\cU}^{vv}}^{-1} )\big(\bS_r+\text{vec}(\bmu_{r})\text{vec}(\bmu_{r})^{\T}\big)\Big)  \notag \\ + K \log |\bSigma_{\cX}| + n \log |\bSigma_{\cU}^{vv}| - \sum_{t=1}^{n} \log | \bS_{tr}|-nK,
\end{gathered}
\end{equation}
% -5pt
where $\bmu_{r} = (\bmu_{1r}, \dots, \bmu_{nr})$, $\bS_r=\text{diag}(\bS_{1r}, \dots, \bS_{nr})$, and 
$%%$
\bSigma_{\cU}^{vv} = \big[\kappa_{\cU}(v_j, v_j)\big]_{1 \leq i, j \leq K}.%\begin{bmatrix}
    %\kappa_{\cU}(v_1, v_1)&   \cdots & \kappa_{\cU}(v_1,v_K)   \\
    %\cdots& \cdots & \cdots \cdots   \\
    %\kappa_{\cU}(v_K,v_1) & \cdots & %\kappa_{\cU}(v_K,v_K)   \\
    %\end{bmatrix}.
$ %$ 
See Appendix~\ref{app:deri_for_divergence} for the detailed derivation. 

\subsection{Sampling for Variational Distribution of Factors}
To optimize the variational distributions, the automatic differentiation variational inference (ADVI) algorithm \cite[]{kucukelbir2017automatic, blei2017, ranganath2014} is adopted to maximize the ELBO in equation~\eqref{eq:elbo_rewrite}. 

To perform ADVI in our model, we need to sample $\bX_r(\cdot)$ from its variational distribution as specified in equation~\eqref{eq:variatonal_conditional}. However, directly sampling from a $nL \times nL$ matrix is computationally expensive even though this distribution is Gaussian conditional on $\bX_r(\bv)$. To address this issue  and accelerate the computation of the ELBO, we take advantage of the separability of the temporal and spatial kernels as described in Section~\ref{subsec:gpdmfd}, and propose the following method. 

For any $\bu = (u_1, \dots, u_L)^{\T}$ with $u_1, \dots, u_L \in \cU$ being the observation points in $\cU$, we first partition the spatial covariance matrix for $\bX(\bu, \bv)$ into a blockwise matrix shown as
$
\begin{bmatrix}
\bSigma_{\cU}^{uu} &{\bSigma_{\cU}^{uv}} \\ 
{\bSigma_{\cU}^{uv}}^{\T} & \bSigma_{\cU}^{vv}
\end{bmatrix},$
where %$\bSigma_{vv}$ is defined above,
$%$
\bSigma_{\cU}^{uu}= \big[\kappa_{\cU}(u_i, u_j)\big]_{1 \leq i, j \leq L},%\begin{bmatrix}
    %\kappa_{\cU}(u_1, u_1)&   \cdots & %\kappa_{\cU}(u_1,u_L)   \\
    %\cdots& \cdots & \cdots \cdots   \\
    %\kappa_{\cU}(u_L,u_1) & \cdots & %\kappa_{\cU}(u_L,u_L)   \\
    %\end{bmatrix}, \ 
%\text{and} \ 
$ and 
$
\bSigma_{\cU}^{uv}= \big[ \kappa_{\cU}(u_i, v_j)\big]_{1 \leq i \leq L, 1 \leq j \leq K}. %\begin{bmatrix}
    %\kappa_{\cU}(u_1, v_1)&   \cdots & %\kappa_{\cU}(u_1,v_K)   \\
    %\cdots& \cdots & \cdots \cdots   \\
    %\kappa_{\cU}(u_L,v_1) & \cdots & %\kappa_{\cU}(u_L,v_K)   \\
    %\end{bmatrix}.
$%$
\begin{theorem}
%[Cancellation of Inter-Task Transfer] 
[Posterior Mean]
The mean function of the posterior for $X_{tr}(\cdot)$ is solely dependent on the variational mean of $\bX_{tr}(\bv)$, the inducing variables at time $t$. That is, for any $\bu$
%%{-5pt}
\label{prop:prop_1}
\begin{equation}
\operatorname{E}\big(X_{tr}(\bu)\big)  
=\bSigma_{\cU}^{uv}{\bSigma_{\cU}^{vv}}^{-1}\bmu_{tr}.
\end{equation}
\end{theorem}
% -5pt
It means that for MTGP, the variational mean is independent of the inducing variables at timestamps other than the current one. See also an analogous theorem for Gaussian process regression 
in \cite{bonilla2007}.
\begin{theorem}
%[Irrelevance of Inter-Task Covariance Matrix] 
[Posterior Variance]
The variance function of the posterior for $\bX_r(\cdot)$ contains two parts. For any $\bu$,
\label{prop:prop_2}
\begin{equation}
\begin{gathered}
\operatorname{Var}_q\big[\operatorname{vec}\big(\bX_{r}(\bu)\big)\big] \!=\!(I \!\otimes\! \bSigma_{\cU}^{uv}{\bSigma_{\cX}^{vv}}^{-1} ) \operatorname{diag}(\bS_{1r}, \dots, \bS_{nr})\\ + \bSigma_{\cX}\otimes   (\bSigma_{\cU}^{uu} - \bSigma_{\cU}^{uv}{\bSigma_{\cU}^{vv}}^{-1}{\bSigma_{\cU}^{uv}}^{\T}).
\end{gathered}
\end{equation}
\end{theorem}
% -5pt
The first part is solely dependent on the variational variance of $\bX_{tr}(\bv)$, while the second part is independent of the variational distributions of all inducing variables.
In particular, the first part corresponds to a group of independent Gaussian processes such that $\tilde{\bX}_{tr}^{(1)}(\bu) \sim \text{Normal}(\bSigma_{\cU}^{uv}{\bSigma_{\cX}^{vv}}^{-1}\bmu_{tr}, \bSigma_{\cU}^{uv}{\bSigma_{\cX}^{vv}}^{-1}\bS_{tr})$ for any $\bu$. One the other hand,  
$\tilde{\bX}_r^{(2)}(\cdot)$ is a zero-mean multi-task Gaussian process, with  $\tilde{\bX}_{r}^{(2)}(\bu) \sim \text{MatrixNormal}(\boldsymbol{0}, 
\bSigma_{\cU}^{uu} - \bSigma_{\cU}^{uv}{\bSigma_{\cU}^{vv}}^{-1}{\bSigma_{\cU}^{uv}}^{\T},
\bSigma_{\cX})$ for any $\bu$.  Therefore, based on Theorems~\ref{prop:prop_1} and \ref{prop:prop_2}, we can decompose
$
{\bX}_r(\cdot) = 
\tilde{\bX}_r^{(1)}(\cdot) + \tilde{\bX}_r^{(2)}(\cdot)
$
under the variational distribution.  Notably, the sampling of 
$\tilde{\bX}_r^{(1)}(\cdot)$
is more efficient as it only depends on inducting variables within the same period.
\begin{theorem}
%[Irrelevance of Inter-Task Covariance Matrix] 
[Irrelevance to ELBO]
Conditional on $\bSigma_{\cX}$ and $\bSigma_{\cU}$, sampling $\bX_{tr}(\cdot)$ from the distribution of $\tilde{\bX}_r^{(1)}(\cdot)$ does not change the variational mean. Moreover, the corresponding ELBO of DF\textsuperscript{2}M  in equation~\eqref{eq:elbo_rewrite} is only modified by a constant term given by 
%% -20pt
\begin{equation}  
\frac{1}{2\sigma_{\epsilon}^2} 
\left \|\bZ \odot \bA\right \|_F^2\operatorname{trace}\big[\bSigma_{\cX}\big] \operatorname{trace}\big[\bSigma_{\cU}^{uu} - \bSigma_{\cU}^{uv}{\bSigma_{\cU}^{vv}}^{-1}{\bSigma_{\cU}^{uv}}^{\T}\big], \end{equation}
\label{prop:prop_3}
where $\|{\bf M}\|_F= \big(\sum_{i,j} M_{ij}^2\big)^{\frac{1}{2}}$ denotes the Frobenius norm of any matrix ${\bf M}$.
\end{theorem}
See Appendices~\ref{app:prof_for_prop_1}, \ref{app:prof_prop_2} and \ref{app:prof_prop_3} for the derivations of Theorem~\ref{prop:prop_1}, \ref{prop:prop_2} and \ref{prop:prop_3}, respectively. With the help of these theorems, we can sample  $\bX_{tr}(\cdot)$ from the proxy variational distribution $\text{Normal}(\bSigma_{\cU}^{uv}{\bSigma_{\cX}^{vv}}^{-1}\bmu_{tr}, \bSigma_{\cU}^{uv}{\bSigma_{\cX}^{vv}}^{-1}\bS_{tr})$, which relies solely on the variational distributions at time $t$. This approach provides a more efficient way of computing the ELBO compared to direct sampling, which requires the complete Cholesky decomposition of the $nL \times nL$ matrix.

\subsection{Initialization, Training and Prediction}
%Functional principal component analysis (FPCA) or dynamic FPCA \cite[]{bathia2010, hormann2015} can be employed to obtain initial values for the factors. The number of factors is predetermined at the start of the model, with the option to adjust them later using Bayesian nonparametrics during training. The deep learning portion of the model is trained using mean squared error (MSE) to achieve accurate initial values for further training. 
We use the technique of ADVI to train the variational parameters of the posteriors, by computing the gradient of the ELBO with respect to the parameters. 
The training process requires iterating through the following steps until the ELBO converges. The steps of Bayesian inference for DF\textsuperscript{2}M are summarized in Algorithm~\ref{alg:alg1} in Appendix~\ref{app:algo}.

First, conditional on $\bSigma_{\cX}$, we update the variational distribution parameters $\bmu_{tr}$ and $\bS_{tr}$ for inducing variables $\bX_{tr}(\bv)$ for all $t$ and $r$, as well as other variational parameters including $\{\tau_j^1, \tau_j^2\}_{1\leq j \leq M}$ and $\{m_{tj}\}_{1 \leq t \leq n, 1\leq j \leq M}$ for India buffet process $\bZ$, $\{\eta_{tj}, \sigma_{tj}^A\}_{1 \leq t \leq n, 1\leq j \leq M}$ for loading weight matrix $\bA$. We also update the idiosyncratic noise scale $\sigma_\epsilon$ and the parameters in the spatial kernel $\kappa_{\cU}(\cdot, \cdot)$. In this step, the gradient of ELBO is accelerated by sampling $X_{tr}(\cdot)$ independently according to Theorem~\ref{prop:prop_3} and the analytical expression for the KL divergence in equation~\eqref{eq:divergence}.

Second, conditional on a sample of $\bX_r(\cdot)$, we update the trainable parameters in sequential deep learning framework $H$ that constructs the temporal kernels $\kappa_{\cX}(\cdot, \cdot)$, via the gradient of ELBO with respect to $\bSigma_{\cX}$. Although any mapping function $F$ can be used in our model, it is natural to choose $F\big(\bX_{t}(\cdot)\big)=\bX_t(\bv)$, which eliminates the need to sample $\bX_r(\cdot)$ when computing the gradient. This is inspired by the fact that the variational distribution of inducing variables can be regarded as sufficient statistics of the Gaussian processes \cite[]{titsias2009}.

Once we have observed the data at time $n$, we use the trained model to generate a posterior distribution that captures our updated understanding of the underlying patterns in the data. Based on this distribution, we make a prediction for the value of the data at the next time step, $n+1$. We present the one-step ahead prediction as:
% -10pt
\begin{equation}
\begin{gathered}
\label{eq:prediction_x}
\bar{\bY}_{n+1}(\bu)=(\bar{\bZ} \odot \bar{\bA}) \bar{\bX}_{n+1}(\bu),\\  \bar{X}_{n+1,r}(\bu) = \bSigma_{\cU}^{uv}{\bSigma_{\cU}^{vv}}^{-1} \bmu_{r}
\bSigma_{\cX}^{-1} {\bSigma_{\cX}^{n+1, 1:n}}^{\T},
\end{gathered}
\end{equation}
% -10pt
where $\bar{\bY}$ and $\bar{\bX}$ represent the predictive means for the observations and factors, respectively. The terms $\bar{\bZ}$ and $\bar{\bA}$ are the posterior means of $\bZ$ and $\bA$, respectively. The component $\bSigma_{\cX}^{n+1, 1:n}$ is a $1 \times n$ matrix given by $\bSigma_{\cX}^{n+1, 1:n}= \big[
\kappa_{\cX}(\cX_{n+1}, \cX_{0}),
\cdots , \kappa_{\cX}(\cX_{n+1},\cX_{n})
\big]. $
See Appendix~\ref{app:deri_for_prediction} for the derivations.
By repeating this process iteratively, we can generate a sequence of predictions for future time steps, thereby forecasting the behavior of the system over time.
%\subsection{Initialization}

%\section{Comparison to Other Methods}
%To be added.

\begin{figure*}[ht]
\centering
\scalebox{0.88}{
%\vspace{-20pt}
\begin{minipage}[b]{0.075\textwidth}
    \centering
\raisebox{0.9cm}{
\rotatebox{90}{\begin{minipage}{2.08cm} \centering{\fontsize{8}{10}\selectfont (1) $\bY_{tj}(\cdot)$ \\  Functional time \\ series observations}
\end{minipage}}
}
\raisebox{1.0cm}{
\rotatebox{90}{\begin{minipage}{2.1cm} \centering{\fontsize{8}{10}\selectfont (2) $\bX_{tr}(\cdot)$\\  Inferred  \\ functional factors}
\end{minipage}}
}
\raisebox{1.0cm}{
\rotatebox{90}{\begin{minipage}{2.2cm}\centering {\fontsize{8}{10}\selectfont (3) $\bSigma_\cX$\\ Temporal   \\ covariance matrices}
\end{minipage}}
}
\end{minipage}%
\begin{subfigure}[b]{0.245\textwidth}
\centering
    \begin{subfigure}{\textwidth}
    \includegraphics[width=\textwidth]{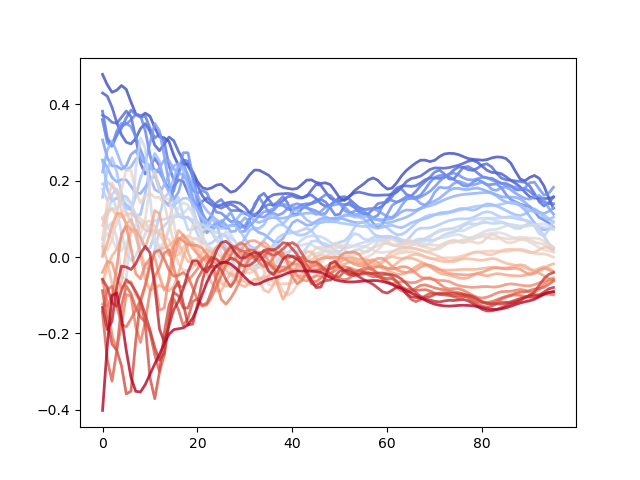}
    \end{subfigure}
     \vskip -6pt
    \begin{subfigure}{\textwidth}
    \includegraphics[width=\textwidth]{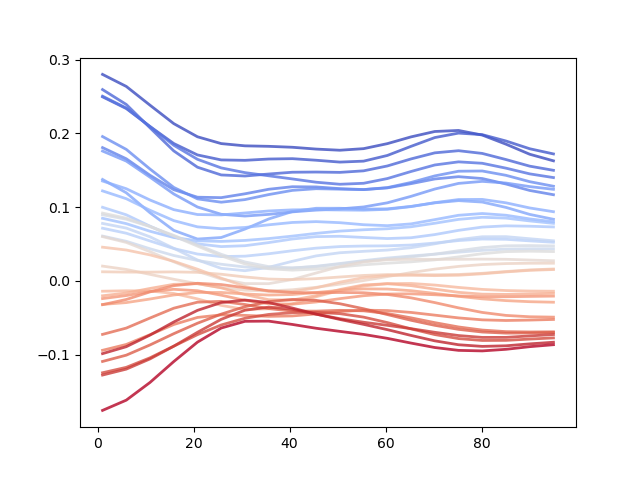}
    \end{subfigure}
    \vskip  -6pt
    \begin{subfigure}{\textwidth}
    \includegraphics[width=1.1\textwidth]{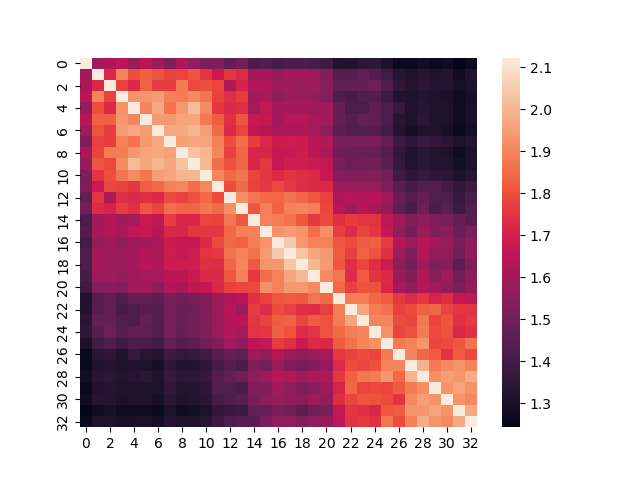}
    \end{subfigure}
    \vskip  -6pt
\centering
\caption{\textit{Japanese mortality}}
\label{fig:fig1a}
\end{subfigure}
\hspace{-10pt}
\begin{subfigure}[b]{0.245\textwidth}
\centering
    \begin{subfigure}{\textwidth}
    \includegraphics[width=\textwidth]{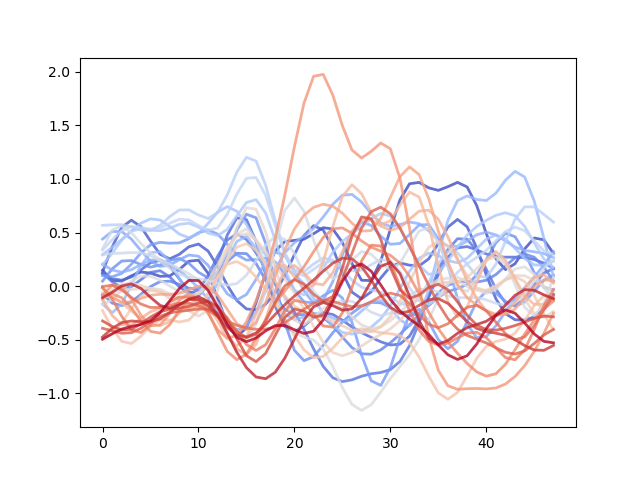}
    \end{subfigure}
        \vskip  -6pt
    \begin{subfigure}{\textwidth}
    \includegraphics[width=\textwidth]{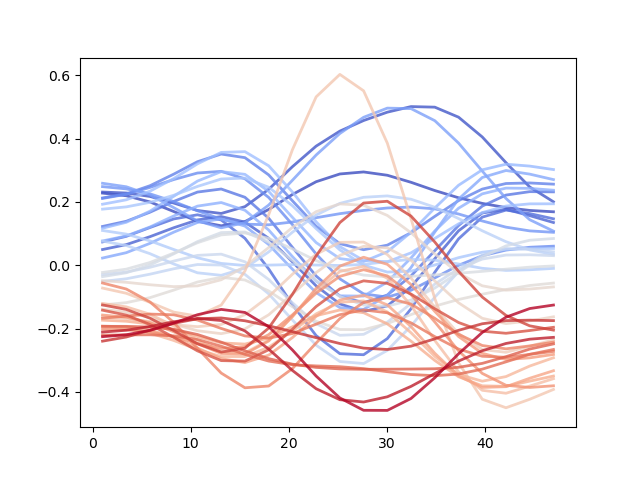}
    \end{subfigure}
        \vskip  -6pt
    \begin{subfigure}{\textwidth}
    \includegraphics[width=1.1\textwidth]{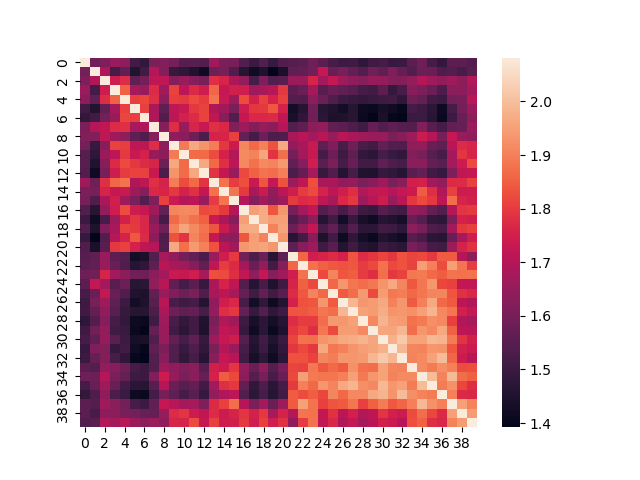}
    \end{subfigure}
        \vskip  -6pt
\centering
\caption{\textit{Energy consumption}}
\label{fig:fig1b}
\end{subfigure}
\hspace{-10pt}
\begin{subfigure}[b]{0.245\textwidth}
\centering
    \begin{subfigure}{\textwidth}
    \includegraphics[width=\textwidth]{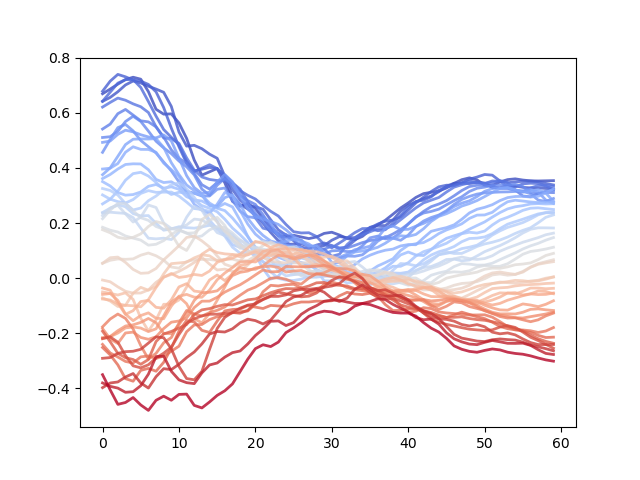}
    \end{subfigure}
        \vskip  -6pt
    \begin{subfigure}{\textwidth}
    \includegraphics[width=\textwidth]{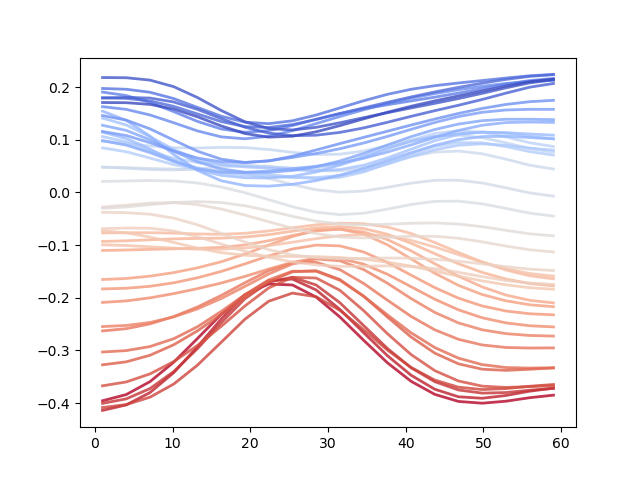}
    \end{subfigure}
        \vskip  -6pt
    \begin{subfigure}{\textwidth}
    \includegraphics[width=1.1\textwidth]{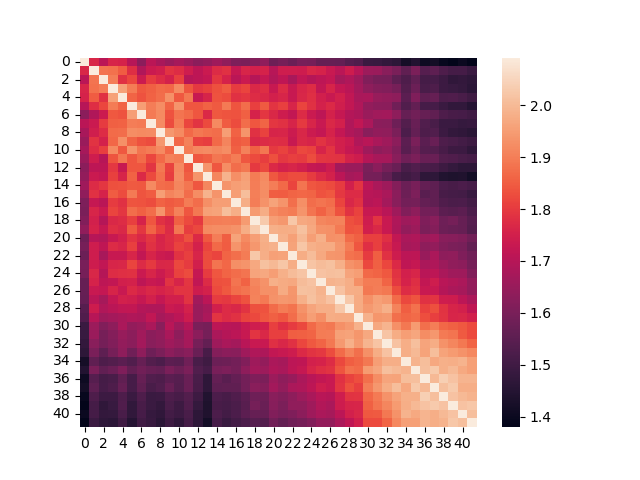}
    \end{subfigure}
        \vskip  -6pt
\centering
\caption{\textit{Global mortality}}
\label{fig:fig1c}
\end{subfigure}
\hspace{-10pt}
\begin{subfigure}[b]{0.245\textwidth}
\centering
    \begin{subfigure}{\textwidth}
    \includegraphics[width=\textwidth]{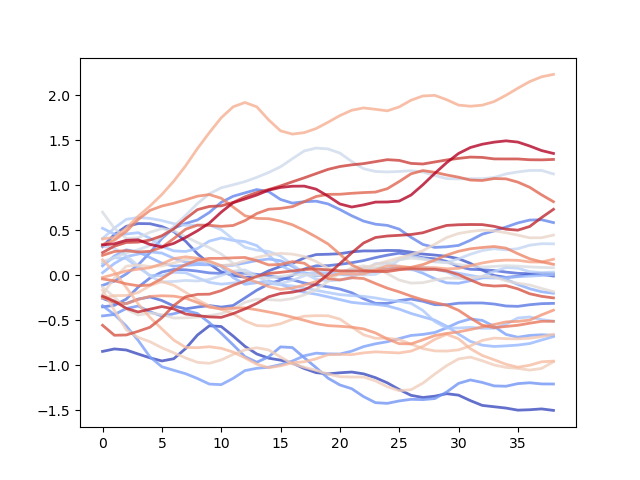}
    \end{subfigure}
        \vskip  -6pt
    \begin{subfigure}{\textwidth}
    \includegraphics[width=\textwidth]{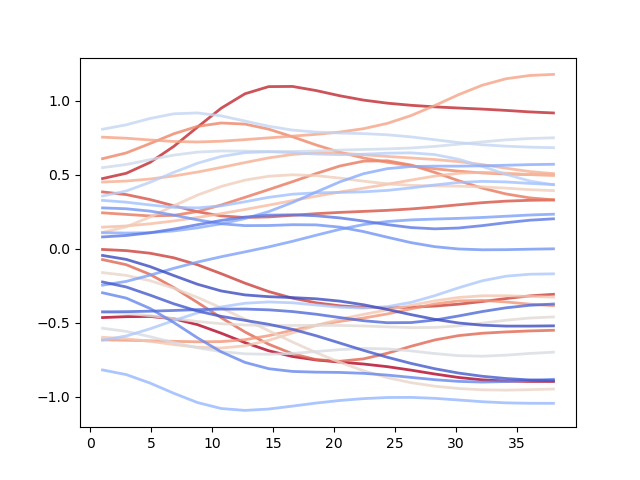}
    \end{subfigure}
        \vskip  -6pt
    \begin{subfigure}{\textwidth}
    \includegraphics[width=1.1\textwidth]{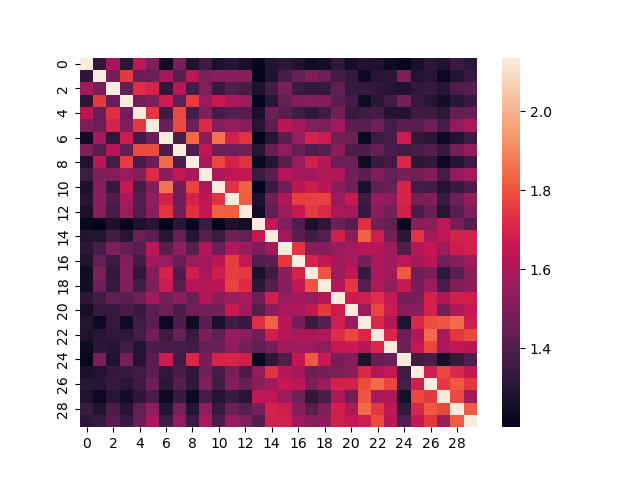}
    \end{subfigure}
        \vskip  -6pt
\centering
\caption{\textit{Stock Intraday}}
\label{fig:fig1d}
\end{subfigure}
}
\caption{A visualization of real datasets with analysis. Row~(1): raw functional time series. Row~(2): the largest functional factor. Row~(3): temporal covariance matrix. 
Rows~(1)~and~(2) use a blue-to-red gradient to denote time progression. Blue for older and red for recent data. Row (3) employs brightness variations to represent covariance, with brighter areas indicating higher covariance.}
%\vskip -10pt
\label{fig:fig1}
\end{figure*}

\section{Experiments}
\subsection{Datasets}
We apply DF\textsuperscript{2}M  to four real-world datasets consisting of high-dimensional functional time series. %, and compare their performance to ordinary deep learning counterparts, including LSTM, GRU and attention mechanism.
{\bf \textit{Japanese Mortality}} dataset contains age-specific  mortality rates for 47 Japanese prefectures ($p$=47) from 1975 to 2017, with 43 observations per prefecture ($n$=43).
{\bf  \textit{Energy Consumption}} dataset includes half-hourly measured energy consumption curves for selected London households ($p$=40) between December 2012 and January 2013 ($n$=55). %The dataset consists of 4000 samples of daily energy consumption curves observed at 48 equally spaced time points. 
{\bf \textit{Global Mortality}} dataset provides a broader perspective on mortality rates by including age-specific mortality data across different countries ($p$=32) from 1960 to 2010 ($n$=50). 
{\bf  \textit{Stock Intraday}} dataset comprises high-frequency price observations for the S\&P 100 component stocks (we removed 2 stocks with missing values, so $p$=98) in 2017. The data includes 45 trading days ($n$=45), with ten-minute resolution prices and cumulative intraday return trajectories \cite[]{horvath2014}. 
Each dataset is preprocessed and transformed into an appropriate format for analysis. See Appendix~\ref{app:data} for the details.
We denote the data as $\big\{Y_{tj}(u_k)\big\}_{1 \leq t \leq n, 1\leq j \leq p, 1 \leq k \leq K}$, where $K$ is the number of observations per curve. Examples of functional time series for a randomly selected $j$ are plotted in Row~(1) of Figure~\ref{fig:fig1}.

Our work is centered on functional time series, distinct from  univariate or multivariate time series. It is important to highlight that functional time series analysis is inherently more complex due to the observations being infinite-dimensional functional objects, making it much more challenging than non-functional time series \cite[]{ramsay2005}.  The choice of datasets with time steps \( n\) deliberately limited to less than 50 is a standard practice in functional time series literature, suitable for demonstrating the robustness of our method in more challenging scenarios with limited data, in line with \cite{chang2023} and \cite{tang2022}.

\begin{table*}[t]
% -5pt
\centering	\caption{Comparision of  DF\textsuperscript{2}M to Standard Deep Learning Models. For formatting reasons, MAPEs are multiplied by 10, and MSPEs are multiplied by 10$^2$, except for the \textit{Energy Consumption} dataset. }
	\label{tab:results}\
%\vskip -7pt
 \centering
\begin{subtable}{0.68\linewidth}
\caption{\scriptsize Comparison of DF\textsuperscript{2}M-LIN and LIN}
\vskip -4pt
\centering
\label{tab:table1a}
\resizebox{1.0\textwidth}{!}{%
  \begin{minipage}{\textwidth}\small
\resizebox{1.0\linewidth}{!}{ 
    \setlength{\tabcolsep}{2.5pt}
    \begin{tabular}{lcccccccccccccccc}
    \toprule
    & & \multicolumn{3}{c}{{\scriptsize Japanese Mortality}} & \multicolumn{3}{c}{{\scriptsize Energy Consumption}} & \multicolumn{3}{c}{{\scriptsize Global Mortality}} & \multicolumn{3}{c}{{\scriptsize Stock Intraday }} \\
    \cmidrule(lr){3-5} \cmidrule(lr){6-8} \cmidrule(lr){9-11} \cmidrule(lr){12-14}
  \ \   $h$ &  & \scriptsize 1 & \scriptsize 2 & \scriptsize 3 & \scriptsize 1 & \scriptsize 2 & \scriptsize 3 & \scriptsize 1 & \scriptsize 2 & \scriptsize 3 & \scriptsize 1 & \scriptsize 2 & \scriptsize 3& \\
    \midrule
        \multirow{2}{*}{\rotatebox{0}{\parbox{0.8cm}{\centering{{\scriptsize DF\textsuperscript{2}M-LIN}} }}} &{\scriptsize MSPE}  & \textbf{4.707}  & \textbf{4.567} & \textbf{5.623} & \textbf{10.29} & \textbf{17.58} & \textbf{17.64} & \textbf{10.78} & \textbf{9.300} & \textbf{9.706} & \textbf{99.58} & \textbf{101.2} & \textbf{89.82} \\
    & {\scriptsize MAPE} & \textbf{1.539} & \textbf{1.446} & \textbf{1.635} & \textbf{2.334} & \textbf{3.060} & \textbf{3.100} & \textbf{2.319} & \textbf{2.041} & \textbf{2.106} & \textbf{6.424} & \textbf{6.505} & \textbf{6.269} \\
    \midrule
    \multirow{2}{*}{\rotatebox{0}{\parbox{0.8cm}{\centering{{\scriptsize LIN}} }}} &{\scriptsize MSPE}  & 7.808 & 8.774 & 9.228 & 16.16 & 18.95 & 20.27 & 16.84 & 18.05 & 19.93 & 137.5 & 127.8 & 139.1 \\
    & {\scriptsize MAPE} & 2.092 & 2.227 & 2.313 & 2.939 & 3.214 & 3.342 & 2.783 & 2.949 & 3.174 & 7.896 & 7.491 & 7.924 \\
    \bottomrule
    \end{tabular}
}     
\end{minipage}}
\end{subtable}
\vskip -3pt
\begin{subtable}{0.66\linewidth}
\caption{\scriptsize Comparison of DF\textsuperscript{2}M-LSTM and LSTM}
\centering
\vskip -4pt
\label{tab:table1b}
\resizebox{1.0\textwidth}{!}{%
  \begin{minipage}{\textwidth}\small
\resizebox{1.0\linewidth}{!}{ 
    \setlength{\tabcolsep}{2.5pt}
    \begin{tabular}{lcccccccccccccccc}
    \toprule
    & & \multicolumn{3}{c}{{\scriptsize Japanese Mortality}} & \multicolumn{3}{c}{{\scriptsize Energy Consumption}} & \multicolumn{3}{c}{{\scriptsize Global Mortality}} & \multicolumn{3}{c}{{\scriptsize Stock Intraday }} \\
    \cmidrule(lr){3-5} \cmidrule(lr){6-8} \cmidrule(lr){9-11} \cmidrule(lr){12-14}
  \ \   $h$ &  & \scriptsize 1 & \scriptsize 2 & \scriptsize 3 & \scriptsize 1 & \scriptsize 2 & \scriptsize 3 & \scriptsize 1 & \scriptsize 2 & \scriptsize 3 & \scriptsize 1 & \scriptsize 2 & \scriptsize 3& \\
    \midrule
        \multirow{2}{*}{\rotatebox{0}{\parbox{0.8cm}{\centering{{\scriptsize DF\textsuperscript{2}M-LSTM}} }}} &{\scriptsize MSPE}  & \textbf{3.753} & \textbf{4.164} & \textbf{4.513} & \textbf{8.928} & \textbf{11.60} & \textbf{17.26} & \textbf{7.672} & \textbf{8.088} & \textbf{8.954} & \textbf{107.5} & \textbf{118.8} & \textbf{113.6} \\
    & {\scriptsize MAPE} & \textbf{1.205} & \textbf{1.322} & \textbf{1.427} & \textbf{2.176} & \textbf{2.478} & \textbf{3.063} & \textbf{1.726} & \textbf{1.823} & \textbf{1.978} & \textbf{6.741} & \textbf{7.141} & \textbf{7.294} \\
    \midrule
    \multirow{2}{*}{\rotatebox{0}{\parbox{0.8cm}{\centering{{\scriptsize LSTM}} }}} &{\scriptsize MSPE}  & 4.989 & 5.597 & 6.501 & 13.51 & 19.71 & 24.61 & 13.28 & 16.29 & 17.08 & 193.3 & 176.0 & 213.8  \\
    & {\scriptsize MAPE} & 1.447 & 1.523 & 1.684 & 2.635 & 3.278 & 3.759 & 2.332 & 2.572 & 2.680 & 9.281 & 9.283 & 10.20 \\
    \bottomrule
    \end{tabular}
}     
\end{minipage}}
\end{subtable}
\vskip -2pt
\begin{subtable}{0.66\linewidth}
\caption{\scriptsize Comparison of DF\textsuperscript{2}M-GRU and GRU}
\centering
\vskip -4pt
\label{tab:table1c}
\resizebox{1.0\textwidth}{!}{%
  \begin{minipage}{\textwidth}\small
\resizebox{1.0\linewidth}{!}{ 
    \setlength{\tabcolsep}{2.5pt}
    \begin{tabular}{lcccccccccccccccc}
    \toprule
    & & \multicolumn{3}{c}{{\scriptsize Japanese Mortality}} & \multicolumn{3}{c}{{\scriptsize Energy Consumption}} & \multicolumn{3}{c}{{\scriptsize Global Mortality}} & \multicolumn{3}{c}{{\scriptsize Stock Intraday }} \\
    \cmidrule(lr){3-5} \cmidrule(lr){6-8} \cmidrule(lr){9-11} \cmidrule(lr){12-14}
  \ \   $h$ &  & \scriptsize 1 & \scriptsize 2 & \scriptsize 3 & \scriptsize 1 & \scriptsize 2 & \scriptsize 3 & \scriptsize 1 & \scriptsize 2 & \scriptsize 3 & \scriptsize 1 & \scriptsize 2 & \scriptsize 3& \\
    \midrule
        \multirow{2}{*}{\rotatebox{0}{\parbox{0.8cm}{\centering{{\scriptsize DF\textsuperscript{2}M-GRU}} }}} &{\scriptsize MSPE}  & \textbf{4.092} & \textbf{4.395} & \textbf{4.898} & \textbf{9.132} & \textbf{8.714} & \textbf{9.730} & \textbf{8.741} & \textbf{8.714} & \textbf{9.730} & \textbf{102.5} & \textbf{117.3} & \textbf{95.49} \\
    & {\scriptsize MAPE} & \textbf{1.318} & \textbf{1.402} & \textbf{1.537} & \textbf{2.204} & \textbf{1.951} & \textbf{2.110} & \textbf{1.967} & \textbf{1.951} & \textbf{2.110} & \textbf{6.675} & \textbf{7.339} & \textbf{6.649} \\
    \midrule
    \multirow{2}{*}{\rotatebox{0}{\parbox{0.8cm}{\centering{{\scriptsize GRU}} }}} &{\scriptsize MSPE}  & 8.800 & 8.552 & 10.41 & 15.55 & 24.02 & 17.53 & 14.12 & 15.33 & 17.53 & 414.0 & 445.9 & 427.2  \\
    & {\scriptsize MAPE} & 1.691 & 1.809 & 1.865 & 2.872 & 3.518 & 2.597 & 2.211 & 2.403 & 2.597 & 14.12 & 14.66 & 14.07 \\
    \bottomrule
    \end{tabular}
}     
\end{minipage}}
\end{subtable}
\vskip -2pt
\begin{subtable}{0.66\linewidth}
\caption{\scriptsize Comparison of DF\textsuperscript{2}M-ATTN and ATTN}
\centering
\vskip -4pt
\label{tab:table1d}
\resizebox{1.0\textwidth}{!}{%
  \begin{minipage}{\textwidth}\small
\resizebox{1.0\linewidth}{!}{ 
    \setlength{\tabcolsep}{2.5pt}
    \begin{tabular}{lcccccccccccccccc}
    \toprule
    & & \multicolumn{3}{c}{{\scriptsize Japanese Mortality}} & \multicolumn{3}{c}{{\scriptsize Energy Consumption}} & \multicolumn{3}{c}{{\scriptsize Global Mortality}} & \multicolumn{3}{c}{{\scriptsize Stock Intraday }} \\
    \cmidrule(lr){3-5} \cmidrule(lr){6-8} \cmidrule(lr){9-11} \cmidrule(lr){12-14}
  \ \   $h$ &  & \scriptsize 1 & \scriptsize 2 & \scriptsize 3 & \scriptsize 1 & \scriptsize 2 & \scriptsize 3 & \scriptsize 1 & \scriptsize 2 & \scriptsize 3 & \scriptsize 1 & \scriptsize 2 & \scriptsize 3& \\
    \midrule
        \multirow{2}{*}{\rotatebox{0}{\parbox{0.8cm}{\centering{{\scriptsize DF\textsuperscript{2}M-ATTN}} }}} &{\scriptsize MSPE}  &  \textbf{3.608} & \textbf{3.839} & \textbf{3.985} & \textbf{14.22} & 18.70 & 19.03 & \textbf{14.22} & \textbf{18.70} & \textbf{19.03} & 104.2 & 103.4 & 93.93 \\
    & {\scriptsize MAPE} & \textbf{1.119} & \textbf{1.203} & \textbf{1.264} & \textbf{2.741} & \textbf{3.141} & \textbf{3.163} & \textbf{2.741} & \textbf{3.141} & \textbf{3.163} & 6.695 & 6.646 & 6.427 \\
    \midrule
    \multirow{2}{*}{\rotatebox{0}{\parbox{0.8cm}{\centering{{\scriptsize ATTN}} }}} &{\scriptsize MSPE}  & 13.44 & 14.85 & 16.17 & 17.03 & \textbf{17.79} & \textbf{18.24} & 39.52 & 41.83 & 43.95 & \textbf{103.4} & \textbf{98.39} & \textbf{91.21}  \\
    & {\scriptsize MAPE} & 3.166 & 3.363 & 3.546 & 3.130 & 3.216 & 3.268 & 5.332 & 5.506 & 5.643 & \textbf{6.579} & \textbf{6.392} & \textbf{6.257} \\
    \bottomrule
    \end{tabular}
}    
 
\end{minipage}}
\end{subtable}
%\vskip -10pt
\end{table*} 

\subsection{Experiment Setup and Metrics}
To assess the predictive accuracy of the proposed model, we split the data into a training set with the first $n_1$ periods and a test set with the last $n_2$ periods. We use the training set to train the parameters in the model following the steps in Section~\ref{sec:inference}. Then for an integer $h>0$, we make the $h$-step-ahead prediction given the fitted model using the first $n_1$ periods. We then repeat this process by moving the training window by one period, refitting the model, and making the $h$-step-ahead prediction. We compute the mean absolute prediction error (MAPE) and 
mean squared prediction error (MSPE) using the following equations:
%% -5pt
\begin{equation*}
\begin{gathered}
\text{MAPE}(h) \!=\! \frac{1}{M}
\sum_{j=1}^{p} \sum_{k=1}^{K} \sum\limits_{\hspace{0.4cm}\mathclap{t=n_1+h}}^{n} 
 \big| \hat{Y}_{tj}(u_k) - Y_{tj}(u_k) \big|, \\ 
\text{MSPE}(h) \!=\! \frac{1}{M}
\sum_{j=1}^{p} \sum_{k=1}^{K} \sum\limits_{\hspace{0.4cm}\mathclap{t=n_1+h}}^{n} 
\big[ \hat{Y}_{tj}(u_k) - Y_{tj}(u_k) \big]^2,
\end{gathered}
\end{equation*}
%% -15pt
where $M=Kp(n_2-h+1)$. 
In our DF\textsuperscript{2}M implementation, we incorporate three cutting-edge deep learning modules: LSTM, GRU, and the self-attention mechanism. In the deep learning modules, we employ a feedforward neural network equipped with ReLU activation functions to map inputs into a designated hidden layer size. Subsequently, the transformed inputs are channeled through a time-invariant full connected neural network, LSTM, GRU, or self-attention mechanisms, denoted as DF\textsuperscript{2}M-LIN,  DF\textsuperscript{2}M-LSTM, DF\textsuperscript{2}M-GRU and DF\textsuperscript{2}M-ATTN, respectively. For DF\textsuperscript{2}M, the outputs of the deep learning modules are passed to the kernel function, while in conventional deep learning, they are converted to outputs via a linear transformation. We evaluate their performance against conventional deep learning models under the same structural setting and regulations. 
The optimal hyperparameters, along with a detailed description of the deep learning architecture, can be found in Appendix~\ref{app:dl_details}. %The proposed inference algorithm is used to infer DF\textsuperscript{2}M and make predictions.

%% -2pt
\subsection{Empirical Results}
Our primary objective is to improve the explainability of deep learning models like RNNs or transformers, while maintaining or enhancing prediction accuracy.

{\bf Explainability} Firstly, Row~(2) of Figure~\ref{fig:fig1}  shows the temporal dynamic of the largest factors in the fitted models. We can observe a decreasing trend over time for the first three datasets. This is particularly valuable as these factors exhibit a clear and smooth dynamic, which can be used to explain the underlying reasons for changes over time and also to make robust predictions. Secondly, the temporal covariance matrix  ($\bSigma_{\cX}$) can be seen in Row~(3) of Figure~\ref{fig:fig1}. It is evident that the first three datasets exhibit stronger autocorrelation than the \textit{Stock Intraday} dataset, which aligns with the intuition that financial data is generally noisier and characterized by short-term dependencies.

Furthermore, both mortality datasets display a strong autoregressive pattern, as evidenced by the large covariance values close to the diagonal. They also show a blockwise pattern, which indicates the existence of change points in 1980s.  Another interesting observation is the periodic pattern in the \textit{Energy Consumption} dataset, which reveals distinct patterns for weekdays and weekends during the first 21 days. This data corresponds to the first 21 days in December. In contrast, the second half of the time steps do not exhibit this pattern. This could be attributed to the Christmas holidays in London, during which the differences between weekdays and weekends are   smaller, as people are on holiday.

%%{-30pt}
{\bf Predictive Accuracy} Compared to standard deep learning models, the DF\textsuperscript{2}M framework consistently outperforms other models in terms of both MSPE and  MAPE across all four datasets. The only exception to this is the \textit{Stock Intraday} dataset, where DF\textsuperscript{2}M-ATTN and ATTN achieve similar levels of accuracy. Specifically, the DF\textsuperscript{2}M-LSTM model performs exceptionally well on the \textit{Energy Consumption} and \textit{Global Mortality} datasets, while the DF\textsuperscript{2}M-ATTN model exhibits the lowest prediction error for the \textit{Japanese Mortality} dataset. These results demonstrate that the integration of an explainable structure with the nonlinearity of LSTM and attention mechanisms can significantly improve the overall performance of the model.

On the other hand, the DF\textsuperscript{2}M-LIN model outperforms both DF\textsuperscript{2}M-LSTM and DF\textsuperscript{2}M-GRU on the \textit{Stock Intraday} dataset. This can be attributed to the fact that, in the context of financial data, long-term dependencies may not be present, rendering the Markovian model more suitable for capturing the underlying dynamics. Consequently, the DF\textsuperscript{2}M-LIN model emerges as a better choice for the \textit{Stock Intraday} dataset. Compared to  standard deep learning models with multiple layers, DF\textsuperscript{2}M achieves better or comparable results, as shown in Appendix~\ref{app:dl_details}. However, in such cases, standard deep learning models sacrifice explainability due to their utilization of a large number of layers.

%%{-2pt}
\section{Related Works}
In the literature concerning frequentist statistical methods for high-dimensional functional time series, various approaches have been employed. For instance, principal components based dimension reduction \cite[]{guoqiao2023,chang2023}, factor model \cite[]{guo2021a} and segmentation transformation \cite[]{chang_qiao_fang_yao_2023}. However, all these methods use either  vector autoregressive (VAR) or functional VAR to describe the temporal dynamics, implying linear and Markovian models. In contrast, our work is the first to propose a Bayesian model for high-dimensional functional time series that can handle nonlinear and non-Markovian dynamics.

Moreover, several studies \cite[]{wilson2016, al2017, xue2019, li2019, watson2021,fortuin2022} have utilized deep kernels in Gaussian processes for classification or regression tasks. In contrast, our framework introduces the use of a deep kernel specifically designed for time series prediction. 

Furthermore, prior work \cite[]{lawrence2003, wang2005, titsias2010} has employed MTGPs to model cross-sectional correlations among  static data. In contrast, we propose the use of a factor model to describe cross-sectional relationships, where the temporal kernel is constructed based on the features of the historical functional factors. This unique structure represents a novel contribution to the current literature.

%% -2pt
\section{Conclusion}
\label{sec:conclusion}
In this paper, we present DF\textsuperscript{2}M, a novel deep Bayesian nonparametric approach for discovering non-Markovian and nonlinear dynamics in high-dimensional functional time series. DF\textsuperscript{2}M combines the strengths of the Indian buffet process, factor model,  Gaussian process, and deep neural networks to offer a flexible and powerful framework. Our model effectively captures non-Markovian and nonlinear dynamics while using deep learning in a structured and explainable manner.  It bridges modern deep learning and statistical time series. We also propose a computationally efficient inference algorithm. 

Empirical results show the superior predictive performance of DF\textsuperscript{2}M compared to corresponding standard deep learning models. However, a potential limitation of our study is its reliance on simple spatial kernels, thereby neglecting to account for the intricate relationships within the observation space. We leave this as an area for future research.

\section*{Acknowledgements}
%Yirui Liu completed this work in the London School of Economics.

This paper was prepared for informational purposes by the CDAO group of JPMorgan Chase \& Co and its affiliates (``J.P. Morgan'') and is not a product of the Research Department of J.P. Morgan. J.P. Morgan makes no representation and warranty whatsoever and disclaims all liability, for the completeness, accuracy or reliability of the information contained herein. This document is not intended as investment research or investment advice, or a recommendation, offer or solicitation for the purchase or sale of any security, financial instrument, financial product or service, or to be used in any way for evaluating the merits of participating in any transaction, and shall not constitute a solicitation under any jurisdiction or to any person, if such solicitation under such jurisdiction or to such person would be unlawful.

%\clearpage
\section*{Impact Statement}
This paper presents work whose goal is to advance the field of Machine Learning. There is no societal consequence of our work.

\bibliographystyle{icml2024}
\bibliography{references_zetora}

\begin{thebibliography}{48}
\providecommand{\natexlab}[1]{#1}
\providecommand{\url}[1]{\texttt{#1}}
\expandafter\ifx\csname urlstyle\endcsname\relax
  \providecommand{\doi}[1]{doi: #1}\else
  \providecommand{\doi}{doi: \begingroup \urlstyle{rm}\Url}\fi

\bibitem[Al-Shedivat et~al.(2017)Al-Shedivat, Wilson, Saatchi, Hu, and
  Xing]{al2017}
Al-Shedivat, M., Wilson, A.~G., Saatchi, Y., Hu, Z., and Xing, E.~P.
\newblock Learning scalable deep kernels with recurrent structure.
\newblock \emph{The Journal of Machine Learning Research}, 18\penalty0
  (1):\penalty0 2850--2886, 2017.

\bibitem[Bathia et~al.(2010)Bathia, Yao, and Ziegelmann]{bathia2010}
Bathia, N., Yao, Q., and Ziegelmann, F.
\newblock Identifying the finite dimensionality of curve time series.
\newblock \emph{The Annals of Statistics}, 38:\penalty0 3352--3386, 2010.

\bibitem[Blei et~al.(2017)Blei, Kucukelbir, and McAuliffe]{blei2017}
Blei, D.~M., Kucukelbir, A., and McAuliffe, J.~D.
\newblock Variational inference: a review for statisticians.
\newblock \emph{Journal of the American Statistical Association}, 112\penalty0
  (518):\penalty0 859--877, 2017.

\bibitem[Bonilla et~al.(2007)Bonilla, Chai, and Williams]{bonilla2007}
Bonilla, E.~V., Chai, K., and Williams, C.
\newblock Multi-task gaussian process prediction.
\newblock In \emph{Advances in {Neural} {Information} {Processing} {Systems}},
  volume~20, 2007.

\bibitem[Chang et~al.(2023{\natexlab{a}})Chang, Chen, Qiao, and Yao]{chang2023}
Chang, J., Chen, C., Qiao, X., and Yao, Q.
\newblock An autocovariance-based learning framework for high-dimensional
  functional time series.
\newblock \emph{Journal of Econometrics}, 2023{\natexlab{a}}.

\bibitem[Chang et~al.(2023{\natexlab{b}})Chang, Fang, Qiao, and
  Yao]{chang_qiao_fang_yao_2023}
Chang, J., Fang, Q., Qiao, X., and Yao, Q.
\newblock On the modelling and prediction of high-dimensional functional time
  series.
\newblock \emph{Working Paper}, 2023{\natexlab{b}}.

\bibitem[Chen et~al.(2022)Chen, Guo, and Qiao]{chen2022}
Chen, C., Guo, S., and Qiao, X.
\newblock Functional linear regression: dependence and error contamination.
\newblock \emph{Journal of Business and Economic Statistics}, 40:\penalty0
  444--457, 2022.

\bibitem[Cho et~al.(2014)Cho, Van~Merriënboer, Bahdanau, and
  Bengio]{cho2014properties}
Cho, K., Van~Merriënboer, B., Bahdanau, D., and Bengio, Y.
\newblock On the properties of neural machine translation: {Encoder}-decoder
  approaches.
\newblock \emph{arXiv preprint arXiv:1409.1259}, 2014.

\bibitem[Damianou \& Lawrence(2013)Damianou and Lawrence]{damianou2013}
Damianou, A. and Lawrence, N.~D.
\newblock Deep {Gaussian} processes.
\newblock In \emph{Proceedings of the {Sixteenth} {International} {Conference}
  on {Artificial} {Intelligence} and {Statistics}}, pp.\  207--215, 2013.

\bibitem[Dawid(1981)]{dawid1981}
Dawid, A.~P.
\newblock Some matrix-variate distribution theory: notational considerations
  and a {Bayesian} application.
\newblock \emph{Biometrika}, 68\penalty0 (1):\penalty0 265--274, 1981.

\bibitem[Fang et~al.(2022)Fang, Guo, and Qiao]{fang2022}
Fang, Q., Guo, S., and Qiao, X.
\newblock Finite sample theory for high-dimensional functional/scalar time
  series with applications.
\newblock \emph{Electronic Journal of Statistics}, 16:\penalty0 527--591, 2022.

\bibitem[Fortuin(2022)]{fortuin2022}
Fortuin, V.
\newblock Priors in {Bayesian} deep learning: a review.
\newblock \emph{International Statistical Review}, pp.\  12502, 2022.

\bibitem[Gao et~al.(2019)Gao, Shang, and Yang]{gao2019}
Gao, Y., Shang, H.~L., and Yang, Y.
\newblock High-dimensional functional time series forecasting: {An} application
  to age-specific mortality rates.
\newblock \emph{Journal of Multivariate Analysis}, 170:\penalty0 232--243,
  2019.

\bibitem[Griffiths \& Ghahramani(2011)Griffiths and Ghahramani]{griffiths2011}
Griffiths, T.~L. and Ghahramani, Z.
\newblock The {Indian} buffet process: an introduction and review.
\newblock \emph{Journal of Machine Learning Research}, 12\penalty0
  (32):\penalty0 1185--1224, 2011.

\bibitem[Guo \& Qiao(2023)Guo and Qiao]{guoqiao2023}
Guo, S. and Qiao, X.
\newblock On consistency and sparsity for high-dimensional functional time
  series with application to autoregressions.
\newblock \emph{Bernoulli}, 29\penalty0 (1):\penalty0 451--472, 2023.

\bibitem[Guo et~al.(2021)Guo, Qiao, and Wang]{guo2021a}
Guo, S., Qiao, X., and Wang, Q.
\newblock Factor modelling for high-dimensional functional time series.
\newblock \emph{arXiv:2112.13651}, 2021.

\bibitem[Guo et~al.(2016)Guo, Liu, Oerlemans, Lao, Wu, and Lew]{guo2016deep}
Guo, Y., Liu, Y., Oerlemans, A., Lao, S., Wu, S., and Lew, M.~S.
\newblock Deep learning for visual understanding: {A} review.
\newblock \emph{Neurocomputing}, 187:\penalty0 27--48, 2016.

\bibitem[Hamelijnck et~al.(2021)Hamelijnck, Wilkinson, Loppi, Solin, and
  Damoulas]{hamelijnck2021}
Hamelijnck, O., Wilkinson, W., Loppi, N., Solin, A., and Damoulas, T.
\newblock Spatio-temporal variational {Gaussian} processes.
\newblock In \emph{Advances in {Neural} {Information} {Processing} {Systems}},
  volume~34, pp.\  23621--23633, 2021.

\bibitem[He et~al.(2016)He, Zhang, Ren, and Sun]{he2016}
He, K., Zhang, X., Ren, S., and Sun, J.
\newblock Deep residual learning for image recognition.
\newblock In \emph{2016 {IEEE} {Conference} on {Computer} {Vision} and
  {Pattern} {Recognition}}, pp.\  770--778, June 2016.

\bibitem[Hochreiter \& Schmidhuber(1997)Hochreiter and
  Schmidhuber]{hochreiter1997long}
Hochreiter, S. and Schmidhuber, J.
\newblock Long short-term memory.
\newblock \emph{Neural computation}, 9\penalty0 (8):\penalty0 1735--1780, 1997.
\newblock Publisher: MIT press.

\bibitem[Hofmann et~al.(2008)Hofmann, Schölkopf, and Smola]{hofmann2008}
Hofmann, T., Schölkopf, B., and Smola, A.~J.
\newblock Kernel methods in machine learning.
\newblock \emph{The annals of statistics}, 36\penalty0 (3):\penalty0
  1171--1220, 2008.

\bibitem[Hormann et~al.(2015)Hormann, Kidzinski, and Hallin]{hormann2015}
Hormann, S., Kidzinski, L., and Hallin, M.
\newblock Dynamic functional principal components.
\newblock \emph{Journal of the Royal Statistical Society: Series B},
  77:\penalty0 319--348, 2015.

\bibitem[Horváth et~al.(2014)Horváth, Kokoszka, and Rice]{horvath2014}
Horváth, L., Kokoszka, P., and Rice, G.
\newblock Testing stationarity of functional time series.
\newblock \emph{Journal of Econometrics}, 179\penalty0 (1):\penalty0 66--82,
  2014.

\bibitem[Hughes \& Sudderth(2013)Hughes and Sudderth]{hughes_memoized_2013}
Hughes, M.~C. and Sudderth, E.
\newblock Memoized online variational inference for {Dirichlet} process mixture
  models.
\newblock In \emph{Advances in {Neural} {Information} {Processing} {Systems}
  26}, pp.\  1133--1141, 2013.

\bibitem[Kucukelbir et~al.(2017)Kucukelbir, Tran, Ranganath, Gelman, and
  Blei]{kucukelbir2017automatic}
Kucukelbir, A., Tran, D., Ranganath, R., Gelman, A., and Blei, D.~M.
\newblock Automatic differentiation variational inference.
\newblock \emph{Journal of machine learning research}, 2017.

\bibitem[Lawrence(2003)]{lawrence2003}
Lawrence, N.
\newblock Gaussian process latent variable models for visualisation of high
  dimensional data.
\newblock \emph{Advances in Neural Information Processing Systems}, 16, 2003.

\bibitem[Li et~al.(2019)Li, Sutherland, Strathmann, and Gretton]{li2019}
Li, W., Sutherland, D.~J., Strathmann, H., and Gretton, A.
\newblock Learning deep kernels for exponential family densities.
\newblock In \emph{International {Conference} on {Machine} {Learning}}, pp.\
  6737--6746, 2019.

\bibitem[Lim \& Zohren(2021)Lim and Zohren]{lim2021}
Lim, B. and Zohren, S.
\newblock Time-series forecasting with deep learning: a survey.
\newblock \emph{Philosophical Transactions of the Royal Society A},
  379\penalty0 (2194):\penalty0 20200209, 2021.
\newblock Publisher: The Royal Society Publishing.

\bibitem[Liu et~al.(2022)Liu, Qiao, and Lam]{liu2020}
Liu, Y., Qiao, X., and Lam, J.
\newblock {CATVI}: {Conditional} and adaptively truncated variational inference
  for hierarchical bayesian nonparametric models.
\newblock In \emph{Proceedings of the 25th {International} {Conference} on
  {Artificial} {Intelligence} and {Statistics}}, pp.\  3647--3662. PMLR, 2022.

\bibitem[Liu et~al.(2023)Liu, Qiao, Wang, and Lam]{eegnn2023}
Liu, Y., Qiao, X., Wang, L., and Lam, J.
\newblock {EEGNN}: {Edge} enhanced graph neural network with a {Bayesian}
  nonparametric graph model.
\newblock In \emph{Proceedings of {The} 26th {International} {Conference} on
  {Artificial} {Intelligence} and {Statistics}}, pp.\  2132--2146. PMLR, 2023.

\bibitem[Miyato et~al.(2018)Miyato, Kataoka, Koyama, and Yoshida]{miyato2018}
Miyato, T., Kataoka, T., Koyama, M., and Yoshida, Y.
\newblock Spectral normalization for generative adversarial networks.
\newblock In \emph{International {Conference} on {Learning} {Representations}},
  2018.

\bibitem[Moreno-Muñoz et~al.(2018)Moreno-Muñoz, Artés, and
  Álvarez]{moreno2018}
Moreno-Muñoz, P., Artés, A., and Álvarez, M.
\newblock Heterogeneous multi-output gaussian process prediction.
\newblock In \emph{Advances in {Neural} {Information} {Processing} {Systems}},
  volume~31, 2018.

\bibitem[Ramsay \& Silverman(2005)Ramsay and Silverman]{ramsay2005}
Ramsay, J.~O. and Silverman, B.~W.
\newblock \emph{Functional data analysis}.
\newblock Springer, New York, 2005.

\bibitem[Ranganath et~al.(2014)Ranganath, Gerrish, and Blei]{ranganath2014}
Ranganath, R., Gerrish, S., and Blei, D.
\newblock Black box variational inference.
\newblock In \emph{Artificial {Intelligence} and {Statistics}}, pp.\  814--822,
  2014.

\bibitem[Tang et~al.(2022)Tang, Shang, and Yang]{tang2022}
Tang, C., Shang, H.~L., and Yang, Y.
\newblock Clustering and forecasting multiple functional time series.
\newblock \emph{The Annals of Applied Statistics}, 16\penalty0 (4):\penalty0
  2523--2553, December 2022.

\bibitem[Teh et~al.(2006)Teh, Jordan, Beal, and Blei]{teh2006}
Teh, Y.~W., Jordan, M.~I., Beal, M.~J., and Blei, D.~M.
\newblock Hierarchical {Dirichlet} processes.
\newblock \emph{Journal of the American Statistical Association}, 101\penalty0
  (476):\penalty0 1566--1581, 2006.

\bibitem[Titsias(2009)]{titsias2009}
Titsias, M.
\newblock Variational learning of inducing variables in sparse {Gaussian}
  processes.
\newblock In \emph{Proceedings of the 12th {International} {Conference} on
  {Artificial} {Intelligence} and {Statistics}}, pp.\  567--574, April 2009.
\newblock ISSN: 1938-7228.

\bibitem[Titsias \& Lawrence(2010)Titsias and Lawrence]{titsias2010}
Titsias, M. and Lawrence, N.~D.
\newblock Bayesian {Gaussian} process latent variable model.
\newblock In \emph{Proceedings of the 13th international conference on
  artificial intelligence and statistics}, pp.\  844--851, 2010.

\bibitem[Torfi et~al.(2020)Torfi, Shirvani, Keneshloo, Tavaf, and
  Fox]{torfi2020}
Torfi, A., Shirvani, R.~A., Keneshloo, Y., Tavaf, N., and Fox, E.~A.
\newblock Natural language processing advancements by deep learning: {A}
  survey.
\newblock \emph{arXiv preprint arXiv:2003.01200}, 2020.

\bibitem[Vaswani et~al.(2017)Vaswani, Shazeer, Parmar, Uszkoreit, Jones, Gomez,
  Kaiser, and Polosukhin]{vaswani_attention_2017}
Vaswani, A., Shazeer, N., Parmar, N., Uszkoreit, J., Jones, L., Gomez, A.~N.,
  Kaiser, L., and Polosukhin, I.
\newblock Attention {Is} {All} {You} {Need}.
\newblock In \emph{Advances in {Neural} {Information} {Processing} {Systems}},
  volume~30, 2017.

\bibitem[Vu \& Lei(2013)Vu and Lei]{vu2013}
Vu, V.~Q. and Lei, J.
\newblock Minimax sparse principal subspace estimation in high dimensions.
\newblock \emph{The Annals of Statistics}, 41\penalty0 (6):\penalty0
  2905--2947, 2013.

\bibitem[Wang et~al.(2005)Wang, Hertzmann, and Fleet]{wang2005}
Wang, J., Hertzmann, A., and Fleet, D.~J.
\newblock Gaussian process dynamical models.
\newblock In \emph{Advances in {Neural} {Information} {Processing} {Systems}},
  volume~18, 2005.

\bibitem[Watson et~al.(2021)Watson, Lin, Klink, Pajarinen, and
  Peters]{watson2021}
Watson, J., Lin, J.~A., Klink, P., Pajarinen, J., and Peters, J.
\newblock Latent derivative {Bayesian} last layer networks.
\newblock In \emph{Proceedings of {The} 24th {International} {Conference} on
  {Artificial} {Intelligence} and {Statistics}}, pp.\  1198--1206, 2021.

\bibitem[Williams \& Rasmussen(2006)Williams and Rasmussen]{williams2006}
Williams, C.~K. and Rasmussen, C.~E.
\newblock \emph{Gaussian {Processes} for {Machine} {Learning}}.
\newblock MIT Press Cambridge, 2006.

\bibitem[Wilson et~al.(2016)Wilson, Hu, Salakhutdinov, and Xing]{wilson2016}
Wilson, A.~G., Hu, Z., Salakhutdinov, R., and Xing, E.~P.
\newblock Deep kernel learning.
\newblock In \emph{Proceedings of the 19th {International} {Conference} on
  {Artificial} {Intelligence} and {Statistics}}, pp.\  370--378, 2016.

\bibitem[Xue et~al.(2019)Xue, Wu, and Sun]{xue2019}
Xue, H., Wu, Z.-F., and Sun, W.-X.
\newblock Deep spectral kernel learning.
\newblock In \emph{Proceedings of the 28th {International} {Joint} {Conference}
  on {Artificial} {Intelligence}}, pp.\  4019--4025, 2019.

\bibitem[Yao et~al.(2021)Yao, Mueller, and Wang]{yao2021a}
Yao, J., Mueller, J., and Wang, J.-L.
\newblock Deep learning for functional data analysis with adaptive basis
  layers.
\newblock In \emph{International {Conference} on {Machine} {Learning}}, pp.\
  11898--11908. PMLR, 2021.

\bibitem[Zhou \& Dette(2023)Zhou and Dette]{zhou2023}
Zhou, Z. and Dette, H.
\newblock Statistical inference for high-dimensional panel functional time
  series.
\newblock \emph{Journal of the Royal Statistical Society Series B: Statistical
  Methodology}, 85\penalty0 (2):\penalty0 523--549, 2023.

\end{thebibliography}

\clearpage
\thispagestyle{empty}     \pagestyle{empty}
\onecolumn 

\hsize \textwidth
    \linewidth\hsize \toptitlebar {\centering
        {\Large\bfseries Supplementary Material to  ``Deep Functional Factor Models: Forecasting High-Dimensional Functional Time Series via Bayesian Nonparametric Factorization''% \par
        }}
    \bottomtitlebar
    
\appendix
\pagenumbering{arabic}
\setcounter{page}{1}

%{\bf \Large  Appendix}%: Deep Functional Factor Modelling}

This supplementary material contains a short review of sequential deep learning modules in Appendix~\ref{app:dl_modules}, a description of multi-task Gaussian process and its connection to matrix normal distribution in Appendix~\ref{app:matrix_mormal}, the functional version of Gaussian process dynamical model in Appendix~\ref{app:functional_gpdm}, technical derivations and proofs in Appendix~\ref{app:proof}, the algorithm of inference in Appendix~\ref{app:algo}, datasets and their preprocessing in Appendix~\ref{app:data}, deep learning structures and hyperparameters in training  in Appendix~\ref{app:dl_details}, finally the standard deviation of the results in Appendix~\ref{app:std}.

\section{An Introduction for Sequential Deep Learning Modules}
\label{app:dl_modules}
\subsection{Recurrent Neural Networks}
\label{app:lstm_and_gru}

LSTM and GRU are both types of Recurrent Neural Networks (RNNs). They are designed to address the problem of vanishing gradients of vanilla RNNs and to preserve long-term dependencies in the sequential data.

LSTM, proposed by \cite{hochreiter1997long}, is composed of memory cells and gates that control the flow of information into and out of the memory cells. The standard structure of LSTM is composed of three types of gates: input gate, output gate and forget gate. The input gate controls the flow of new information into the memory cell, the output gate controls the flow of information out of the memory cell, and the forget gate controls the information that is removed from the memory cell. The standard structure of LSTM is defined as follows, 
%\begin{figure}
%\begin{subfigure}
%\centering
%\includestandalone[mode=image]{pictures/lstm_pict}
%\caption{Sparse functional factor model}
%\end{subfigure}
%    \caption{Caption}
%    \label{fig:lstm}
%\end{figure}
\begin{align*}
    &\bdf_t = \sigma(\bW_f [\bh_{t-1}, \bx_t] + \bb_f) \\
    &\bi_t = \sigma(\bW_i [\bh_{t-1}, \bx_t] + \bb_i) \\
    &\tilde{\bc}_t = \tanh(\bW_C [\bh_{t-1}, \bx_t] + \bb_c) \\
    &{\bc}_t = \bdf_t \odot \bc_{t-1} + \bi_t \odot \tilde{\bc}_t \\
    &\bo_t = \sigma(\bW_o [\bh_{t-1}, \bx_t] + \bb_o) \\
    &\bh_t = \bo_t \odot \tanh(\tilde{\bc}_t),
\end{align*}
where $[\bh_{t-1}, \bx_t]$ is the stack of hidden state vector $\bh_{t-1}$ and $\bx_{t}$. $\bdf_t$, $\bi_t$, and  $\bo_t$ are the activation vectors for forget gate,  update gate, and output
gate, respectively.  $\tilde{\bc}_t$ is cell input activation vector, and $c_t$ is cell state vector, $\sigma$ denotes sigmoid function. $\bW $s
and $\bb$s refer to weight matrices and bias vectors to be
estimated.

GRU is a simplified version of LSTM. It has two gates: update gate and reset gate. The update gate controls the flow of new information into the memory cell, while the reset gate controls the flow of information out of the memory cell. The structure of GRU is defined as follows,
\begin{align*}
    &\bz_t = \sigma(\bW_z [\bh_{t-1}, \bx_t] + \bb_z) \\
    &\br_t = \sigma(\bW_r [\bh_{t-1}, \bx_t] + \bb_r) \\
    &\tilde{\bh}_t = \tanh(\bW_h [\br_t \odot \bh_{t-1}, \bx_t] + \bb_h) \\
    &\bh_t = (1 - \bz_t) \odot \bh_{t-1} + \bz_t \odot \tilde{\bh}_t,
\end{align*}
where $\bz_t$ and $\br_t$  are the activation vectors for update gate and reset
gate, respectively, and $\tilde{\bh}_t$ is  cell input activation vector.

\subsection{Attention Mechanism}
\label{app:atention}
Attention mechanism is a deep learning model that is especially effective for sequential data prediction. It allows the model to assign different weights to different parts of the input, rather than treating them all equally. This can improve the model's ability to make predictions by allowing it to focus on the most relevant parts of the input.
The commonly used self-attention mechanism computes a weight for each element of the input, and the final output is a weighted sum of the input elements, where the weights are computed based on a query, a set of key-value pairs and a similarity function such as dot-product or MLP. The structure of standard self-attention mechanism is shown as follows,
\begin{align*}
&\boldsymbol{q}_t = \boldsymbol{x}_t \boldsymbol{W}_Q , \quad
    \boldsymbol{k}_t = \boldsymbol{x}_t \boldsymbol{W}_K ,\quad
    \boldsymbol{v}_t = \boldsymbol{x}_t \boldsymbol{W}_V  \\
    &a_{t, s} = \frac{\exp(\boldsymbol{q}_t \cdot \boldsymbol{k}_s^\top / \sqrt{d_k})}{\sum_{i=1}^T \exp(\boldsymbol{q}_t \cdot \boldsymbol{k}_i^\top / \sqrt{d_k})}, \quad
    \boldsymbol{h}_t = \sum_{s=1}^T a_{t, s} \boldsymbol{v}_s,
\end{align*}
where $\boldsymbol{q}_t$, $\boldsymbol{k}_t$, and $\boldsymbol{v}_t$  are query, key, and value vectors at time step $t$, respectively, $a_{t, s}$ is the attention value between time steps $t$ and $s$, $d_k$ is the dimension of the key vector, and $T$ is the total number of time steps in the sequence.

In particular, for time series modeling, the attention value should only depend on historical information rather than future information. Therefore, the attention value should be revised as
\begin{equation*}
    a_{t, s} = \frac{\exp(\boldsymbol{q}_t \cdot \boldsymbol{k}_s^\top / \sqrt{d_k})\mathds{1}_{s\leq t}}{\sum_{i=1}^T \exp(\boldsymbol{q}_t \cdot \boldsymbol{k}_i^\top / \sqrt{d_k})\mathds{1}_{s\leq t}},
\end{equation*}
\vskip -5pt
We illustrate the differences between RNN and attention mechanisms in Figure~\ref{fig:structures}. In RNNs, the current state depends on the most recent state, implying a sequential dependence on past states. By contrast, attention mechanisms allow the current state to depend directly on all past states, providing a more flexible and potentially more expressive way to capture the relationships between past and current states in the time series.

\renewcommand{\thefigure}{A.\arabic{figure}}
\setcounter{figure}{0}
\begin{figure*}[t!]
\centering
\begin{subfigure}{0.58\textwidth}
\centering
\begin{tikzpicture}[item/.style={draw,thick,align=center},
itemc/.style={item,on chain,join}]
 \begin{scope}[start chain=going right,nodes=itemc,every
 join/.style={-latex,very thick},local bounding box=chain, ]
 \path node[fill=white] (A1) {\makecell[c]{Neural\\  Networks}} 
 node[fill=white] (A2) {\makecell[c]{Neural\\  Networks}}  
 node[xshift=2em, fill=white] (At-1)  {\makecell[c]{Neural\\  Networks}}  
 node[fill=white] (At) {\makecell[c]{Neural\\  Networks}}  
 node[fill=white] (At+1) {\makecell[c]{Neural\\  Networks}}  
 node[xshift=2em, fill=white] (An) {\makecell[c]{Neural\\  Networks}} ;
 \end{scope}
 \node[left=0.5em of chain,scale=2] (eq) {$=$};
 \node[left=0.5em of eq, item, minimum size=3em, fill=white] (AL) {RNN \ Layer};
 \path (AL.west) ++ (-1em,2em) coordinate (aux);
 %\draw[very thick,-latex,rounded corners] (AL.east) -| ++ (1em,2em) -- (aux) |- (AL.west);
 \foreach \X in {1,2,t-1,t, t+1, n} 
 {\draw[very thick,-latex] (A\X.north) -- ++ (0,2em)
 node[circle,above,item,fill=gray!10, minimum size=3em] (h\X) {$\boldsymbol{h}_{\X}$};
 \draw[circle,very thick,latex-] (A\X.south) -- ++ (0,-2em)
 node[below,item,fill=gray!10,minimum size=3em] (x\X) {$\boldsymbol{x}_{\X}$};}
 
 \draw[white,line width=0.8ex] (AL.north) -- ++ (0,1.9em);
 \draw[very thick,-latex] (AL.north) -- ++ (0,2em)
 node[circle,above,item,fill=gray!10, minimum size=3em] {$\boldsymbol{h}_{t}$};
 \draw[very thick,latex-] (AL.south) -- ++ (0,-2em)
 node[circle,below,item,fill=gray!10, minimum size=3em] {$\boldsymbol{x}_{1:t}$};
 \path (x2) -- (xt-1) node[midway,scale=2,font=\bfseries] {\dots};
  \path (xt+1) -- (xn) node[midway,scale=2,font=\bfseries] {\dots};
\end{tikzpicture}
\caption{RNN Layer}
\label{fig:subfig_0}
\end{subfigure}
% 0.5em
\begin{subfigure}{0.58\textwidth}
\centering
\begin{tikzpicture}[item/.style={draw,thick,align=center},
itemc/.style={item,on chain,join}]
 \begin{scope}[start chain=going right,nodes=itemc,every
 join/.style={-latex,very thick},local bounding box=chain, ]
 \path node[fill=white] (A1) {\makecell[c]{\ Query,\ \  \\ Keys, \\ Values}} 
 node[fill=white] (A2) {\makecell[c]{\ Query,\ \  \\Keys,\\ Values}}  
 node[xshift=2em, fill=white] (At-1)  {\makecell[c]{\ Query,\ \  \\Keys,\\Values}}  
 node[fill=white] (At) {\makecell[c]{\ Query,\ \   \\Keys,\\Values}}  
 node[fill=white] (At+1) {\makecell[c]{\ Query,\ \  \\Keys, \\ Values}}  
 node[xshift=2em, fill=white] (An) {\makecell[c]{\ Query,\ \  \\Keys, \\Values}} ;
 \end{scope}
 \node[left=0.5em of chain,scale=2] (eq) {$=$};
 \node[left=0.5em of eq, item, minimum size=3em, fill=white] (AL) {Attention \\ Mechanism};
 \path (AL.west) ++ (-1em,2em) coordinate (aux);
 %\draw[very thick,-latex,rounded corners] (AL.east) -| ++ (1em,2em) -- (aux) |- (AL.west);
 \foreach \X in {1,2,t-1,t, t+1, n} 
 {\draw[very thick,-latex] (A\X.north) -- ++ (0,2.5em)
 node[circle,above,item,fill=gray!10, minimum size=3em] (h\X) {$\bh_{\X}$};
 \draw[circle,very thick,latex-] (A\X.south) -- ++ (0,-2em)
 node[below,item,fill=gray!10,minimum size=3em] (x\X) {$\bx_{\X}$};}
 
 \draw[white,line width=0.8ex] (AL.north) -- ++ (0,1.9em);
 \draw[very thick,-latex] (AL.north) -- ++ (0,2em)
 node[circle,above,item,fill=gray!10, minimum size=3em] {$\bh_{t}$};
 \draw[very thick,latex-] (AL.south) -- ++ (0,-2em)
 node[circle,below,item,fill=gray!10, minimum size=3em] {$\bx_{1:t}$};
 \path (x2) -- (xt-1) node[midway,scale=2,font=\bfseries] {\dots};
  \path (xt+1) -- (xn) node[midway,scale=2,font=\bfseries] {\dots};

\foreach \X in {n}  
{\foreach \Y in {1,2,t-1,t,t+1} {\draw[circle, thick,latex-,color=yellow] (h\X.south) --  (A\Y.north);};}
  
\foreach \X in {t+1}  
  {\foreach \Y in { 1,2,t-1,t} {\draw[circle, thick,latex-,color=orange] (h\X.south) --  (A\Y.north);};}
  
  \foreach \X in {t}  
  {\foreach \Y in {1,2,t-1} {\draw[circle,  thick,latex-,color=green] (h\X.south) --  (A\Y.north);};}
  
 \foreach \X in {t-1}  
{\foreach \Y in {1,2} {\draw[circle, thick,latex-,color=red] (h\X.south) --  (A\Y.north);};}
  
  \foreach \X in {2}  
  {\foreach \Y in {1} {\draw[circle, thick,latex-,color=blue] (h\X.south) --  (A\Y.north);};}
  
   \path (ht+1) -- (hn)
   node[midway] (a){} 
   node[below=0.2em  of a,scale=2,font=\bfseries, color=black] {\dots}; 
  
  \path (h2) -- (ht-1)
   node[midway] (a){} 
   node[below=0.2em  of a,scale=2,font=\bfseries, color=black] {\dots}; 
\end{tikzpicture}
\caption{Attention Mechanism}
\label{fig:subfig_1}
\end{subfigure}
\caption{The structures for sequential deep learning modules. In the attention mechanism, the colored links demonstrate that the current state relies exclusively on past states, ensuring that the model considers historical information without incorporating future data.}
\vskip -10pt
\label{fig:structures}
\end{figure*}

\section{Multi-task Gaussian Process and Matrix Normal Distribution}
\label{app:matrix_mormal}
We first provide a brief introduction of matrix normal distribution.
A random matrix $\boldsymbol{M} \in \R^{m \times n}$ is said to have a matrix normal distribution, denoted as $\boldsymbol{M} \sim \text{MatrixNormal}_{m \times n}(\boldsymbol{M}_0, \boldsymbol{U}, \boldsymbol{V})$, if its probability density function is given by
\begin{equation*}
p(\boldsymbol{M}) = \frac{\exp\left(-\frac{1}{2} \operatorname{trace}\left[\boldsymbol{V}^{-1}(\boldsymbol{M} - \boldsymbol{M}_0)^{\top} \boldsymbol{U}^{-1}(\boldsymbol{M} - \boldsymbol{M}_0)\right]\right)}{(2\pi)^{\frac{mn}{2}} |\boldsymbol{U}|^{\frac{n}{2}} |\boldsymbol{V}|^{\frac{m}{2}}},
\end{equation*}
where $\boldsymbol{M}_0 \in \R^{m \times n}$ is the mean matrix, $\boldsymbol{U} \in \R^{m \times m}$ is a positive definite row covariance matrix, and $\boldsymbol{V} \in \R^{n \times n}$ is a positive definite column covariance matrix.
Moreover,
the distribution of $\operatorname{vec}(\boldsymbol{M})$ is given by
\begin{equation*}
\operatorname{vec}(\boldsymbol{M}) \sim \mathcal{N}_{mn}\left(\operatorname{vec}(\boldsymbol{M}_0), \boldsymbol{V} \otimes \boldsymbol{U}\right),
\end{equation*}
where $\mathcal{N}_{mn}(\cdot, \cdot)$ represents a multivariate normal distribution with dimension $mn$. Here, $\operatorname{vec}(\boldsymbol{M}_0)$ is the mean vector, and the covariance matrix is formed by the Kronecker product of the row covariance matrix $\boldsymbol{V}$ and the column covariance matrix $\boldsymbol{U}$.

For any $u_1, \dots, u_L \in \cU$,  given equation~\eqref{eq:variance_product_model}, we have $\text{vec} \big(\bX_r(u_1, \dots, u_L)\big) \sim \text{Normal}(\boldsymbol{0}, \bSigma_{\cX} \otimes \bSigma_{\cU}^u)$,

where
$$
\bX_r(u_1, \dots, u_L) = 
    \begin{bmatrix}
    X_{1r}(u_1)&   \cdots & X_{nr}(u_1)   \\
    \cdots& \cdots & \cdots \cdots   \\
    X_{1r}(u_L) & \cdots & X_{nr}(u_L)   \\
    \end{bmatrix}, 
$$
$$
\bSigma_{\cX} = \begin{bmatrix}
    \kappa_{\cX}(\cX_{0}, \cX_{0})&   \cdots & \kappa_{\cX}(\cX_{0},\cX_{n-1})   \\
    \cdots& \cdots & \cdots \cdots   \\
    \kappa_{\cX}(\cX_{n-1},\cX_{0}) & \cdots & \kappa_{\cX}(\cX_{n-1},\cX_{n-1})   %\\
   \end{bmatrix}, 
\quad
\text{and} 
\quad
\bSigma_{\cU}^u = \begin{bmatrix}
\kappa_{\cU}(u_1, u_1)&   \cdots & \kappa_{\cU}(u_1,u_L)   \\
    \cdots& \cdots & \cdots \cdots   \\
    \kappa_{\cU}(u_L,u_1) & \cdots & \kappa_{\cU}(u_L,u_L)  % \\
    \end{bmatrix}.
$$
Therefore, $\bX_r(u_1, \dots, u_L) \sim \text{MatrixNormal}(\boldsymbol{0}, \bSigma_{\cU}^u, \bSigma_{\cX})$.

\section{Functional Version of Gaussian Process Dynamical Model}
\label{app:functional_gpdm}
Following \cite{wang2005}, we consider a nonlinear function $g$ with respect to historical information, achieved by a linear combination of nonlinear kernel function $\phi_i$s,
\renewcommand{\theequation}{C.\arabic{equation}} % Change the equation numbering format
\setcounter{equation}{0} % Reset the equation counter
\begin{equation}
\label{eq:linear_comb}
    \bX_t(\cdot) = g(\cX_{t-1}) = \sum_{i} \phi_i(\cX_{t-1}) {\ba}_i(\cdot) ,
\end{equation}
where  $\cX_{t-1} = \{\bX_{t-1}, \bX_{t-2}, \dots\}$ is the set of all historical factors till period $t-1$,  $\phi_i$ is a nonlinear basis function with respect to $\cX_{t-1}$, and $\ba_i(\cdot)$ is a function defined on $\cU$. Equivalently, the equation above can be presented as 
\begin{equation*}
\begin{bmatrix} 
    X_{t1}(\cdot) \\ 
    %X_{t2}(\cdot) \\ 
    \cdots \\ 
    X_{tr}(\cdot) \\ \cdots
    \end{bmatrix}= \sum_{i}  \phi_i(\cX_{t-1})
    \begin{bmatrix} 
    {a}_{1i}(\cdot) \\
    %{a}_{2i}(\cdot) \\
    \cdots \\ 
    {a}_{ri}(\cdot) \\
     \cdots
    \end{bmatrix} ,
\end{equation*}
where $\ba_i(\cdot) = \big\{ {a}_{1i}(\cdot), {a}_{2i}(\cdot), \dots, {a}_{ri}(\cdot),\dots, \big\}^{\T}$.
This functional version of dynamical system corresponds to equation~(3) in \cite{wang2005}. In analogy, the specific form of $g(\cdot)$ in equation~(\ref{eq:linear_comb}), including the numbers of kernel functions, is incidental, and therefore can be marginalized out from a Bayesian perspective. Assigning each $a_{ri}(\cdot)$ an independent  Gaussian process prior with kernel $\kappa_{\cU}$,  marginalizing over $g$ leads to equation~\eqref{eq:variance_product_model}, where $\kappa_{\cX}(\cX_{t-1}, \cX_{s-1}) = \sum_{i} \big\langle \phi_i(\cX_{t-1}), \phi_i(\cX_{s-1}) \big\rangle$.

\section{Technical Derivations and Proofs}
\label{app:proof}
\subsection{Derivations for Equation (\ref{eq:elbo_rewrite})}
\label{app:sparse_variational}
Using the variational seting in equation~\eqref{eq:variatonal_conditional}, the ELBO in equation~\eqref{eq:elbo} can be written as
\begin{align*}
\text{ELBO} = &\E_q \Big[\log p(\bZ\mid \alpha) p(\bA \mid \bSigma_A)  \prod_{t=1}^{n}p\big(\bY_t(\cdot) \mid \bX_t(\cdot), \bZ, \bA\big)\prod_{r\geq1} p\big(\bX_r(\cdot) \mid 
 \kappa_{\cX}, \kappa_{\cU} \big) \Big] \\& 
 %+ \log p(\kappa_{\cX})+ \log p(\kappa_{\cU})
 -   \E_q \Big[\log q(\bZ) q(\bA) \prod_{r\geq1} q\big(\bX_r(\cdot)\big)\Big]\\
 = & \E_q \big[\log p(\bZ\mid \alpha) \big] - \E_q \big[\log q(\bZ) \big] + \E_q \big[\log p(\bA \mid \bSigma_A) \big] - \E_q \big[\log q(\bA) \big] \\
 & +\sum_{t=1}^{n}\E_q \Big[\log  p\big(\bY_t(\cdot) \mid \bX_t(\cdot), \bZ, \bA\big) \Big] \\
 &+ \sum_{r\geq1}\E_q \Big[\log p\big(X_{1r}(\cdot), \dots, X_{nr}(\cdot) \mid X_{1r}(\bv), \dots, X_{nr}(\bv)
    , \kappa_{\cX}, \kappa_{\cU}
    \big) \\
    & \quad \quad \quad \quad \quad \quad  p\big(X_{1r}(\bv), \dots, X_{nr}(\bv) \mid \kappa_{\cX}, \kappa_{\cU}\big) \Big] \\
  &- \sum_{r\geq1}\E_q \Big[\log p\big(X_{1r}(\cdot), \dots, X_{nr}(\cdot) \mid X_{1r}(\bv), \dots, X_{nr}(\bv)
    , \kappa_{\cX}, \kappa_{\cU}
    \big) \\
    & \quad \quad \quad \quad \quad \quad  q\big(X_{1r}(\bv), \dots, X_{nr}(\bv) \big) \Big] .
\end{align*}
Next, we cancel the same items from the equation above to get:
\begin{align*}
\text{ELBO} 
 = & \E_q \big[\log p(\bZ\mid \alpha) \big] - \E_q \big[\log q(\bZ) \big] + \E_q \big[\log p(\bA \mid \bSigma_A) \big] - \E_q \big[\log q(\bA) \big] \\
 & +\sum_{t=1}^{n}\E_q \Big[\log  p\big(\bY_t(\cdot) \mid \bX_t(\cdot), \bZ, \bA\big) \Big] \\
 &+ \sum_{r\geq1}\E_q \Big[\log p\big(X_{1r}(\bv), \dots, X_{nr}(\bv) \mid \kappa_{\cX}, \kappa_{\cU}\big) - \log q\big(X_{1r}(\bv), \dots, X_{nr}(\bv) \big) \Big] .
\end{align*}
Finally, equation~\eqref{eq:elbo_rewrite} is obtained using the definition of KL divergence.

\subsection{Derivations for Equation (\ref{eq:divergence})}
\label{app:deri_for_divergence}
The Kullbuck--Leibler divergence between two $k$-dimensional multivariate Gaussian distribution $\cN_0 = \text{Normal}(\bm_0, \bSigma_0)$ and $\cN_1 = \text{Normal}(\bm_1, \bSigma_1)$ is defined as,
$$
\text{KL}\left({\mathcal {N}}_{0}\parallel {\mathcal {N}}_{1}\right)={\frac {1}{2}}\left(\operatorname {trace} \left(\bSigma _{1}^{-1}\bSigma _{0}\right)-k+\left(\bm _{1}-\bm _{0}\right)^{\mathsf {T}}\bSigma _{1}^{-1}\left(\bm_{1}-\bm _{0}\right)+\log \left({\frac {\det \bSigma _{1}}{\det \bSigma _{0}}}\right)\right).
$$
In our settings, $\bv=(v_1, \dots, v_K)^{\T}$, the prior and variational distributions for $\bX_r(\bv)$ are $$p\Big(\text{vec}\big(\bX_r(\bv)\big)\Big) = \text{Normal}(\boldsymbol{0}, \bSigma_{\cX} \otimes \bSigma_{\cU}^{vv})$$ and $$q\Big(\text{vec}\big(\bX_r(\bv)\big)\Big)=\text{Normal}\Big(\begin{bmatrix}
\bmu_{1r}\\ 
\cdots\\ 
\bmu_{nr}
\end{bmatrix}, \text{diag}(\bS_{1r}, \dots, \bS_{nr})\Big),$$ respectively, where
$$
\bSigma_{\cX} = \begin{bmatrix}
    \kappa_{\cX}(\cX_{0}, \cX_{0})&   \cdots & \kappa_{\cX}(\cX_{0},\cX_{n-1})   \\
    \cdots& \cdots & \cdots \cdots   \\
    \kappa_{\cX}(\cX_{n-1},\cX_{0}) & \cdots & \kappa_{\cX}(\cX_{n-1},\cX_{n-1})   \\
    \end{bmatrix}, 
\quad
\bSigma_{\cU}^{vv} = \begin{bmatrix}
    \kappa_{\cU}(v_1, v_1)&   \cdots & \kappa_{\cU}(v_1,v_K)   \\
    \cdots& \cdots & \cdots \cdots   \\
    \kappa_{\cU}(v_K,v_1) & \cdots & \kappa_{\cU}(v_K,v_K)   \\
    \end{bmatrix}.
$$

Let $\bm_0 =\begin{bmatrix}
\bmu_{1r}\\ 
\cdots\\ 
\bmu_{nr}
\end{bmatrix}$, $\bm_1 = \boldsymbol{0}$, $\bSigma_0 =\text{diag}(\bS_{1r}, \dots, \bS_{nr})$ and $\bSigma_1 =\bSigma_{\cX} \otimes \bSigma_{\cU}^{vv}$, we  have 
$$\text{trace}(\bSigma _{1}^{-1}\bSigma _{0}) = \text{trace}\big((\bSigma_{\cX}^{-1} \otimes {\bSigma_{\cU}^{vv}}^{-1} ) \text{diag}(\bS_{1r}, \dots, \bS_{nr}) \big),$$ $$\text{det}(\bSigma_1) = |\bSigma_{\cX}|^M|\bSigma_{\cU}^{vv}|^n, \quad \text{det}(\bSigma_0) = \prod_{t=1}^n |\bS_{tr}|,$$ and $$\left(\bmu _{1}-\bmu _{0}\right)^{\mathsf {T}}\bSigma _{1}^{-1}\left(\bmu_{1}-\bmu _{0}\right)=\text{trace}(\bSigma _{1}^{-1}\bmu _{0}\bmu _{0}^{\T})=\text{trace}\big((\bSigma_{\cX}^{-1} \otimes {\bSigma_{\cU}^{vv}}^{-1})\text{vec}(\bmu_{r})\text{vec}(\bmu_{r})^{\T}\big).$$

Therefore,
\begin{align*}
2\text{KL} \big(q(\bv_r)&\parallel p(\bv_r \mid \kappa_{\cX}, \kappa_{\cU})
 \big) =\text{trace}\Big((\bSigma_{\cX}^{-1} \otimes {\bSigma_{\cU}^{vv}}^{-1} )\big(\bS_r+\text{vec}(\bmu_{r})\text{vec}(\bmu_{r})^{\T}\big)\Big) \\&+ K \log |\bSigma_{\cX}| + n \log |\bSigma_{\cU}^{vv}| - \sum_{t=1}^{n} \log | \bS_{tr}|-nK,
\end{align*}
where $\bmu_{r} = (\bmu_{1r}, \dots, \bmu_{nr})$ and $\bS_r=\text{diag}(\bS_{1r}, \dots, \bS_{nr})$.
Moreover, to get avoid of large matrix computation, we can further simplify 
$$
\text{trace}\big((\bSigma_{\cX}^{-1} \otimes {\bSigma_{\cU}^{vv}}^{-1} )\bS_r\big)=\sum_{t=1}^{n} {\Sigma_{\cX}^{-1}}_{t,t} \text{trace}({\bSigma_{\cU}^{vv}}^{-1}\bS_{tr}), 
$$
where ${\Sigma_{\cX}^{-1}}_{t,t} $ denotes the $(t,t)$-th entry of ${\bSigma_{\cX}^{-1}}$ %in $t$-th row and $t$-th column, 
and 
$$\text{trace}\big((\bSigma_{\cX}^{-1} \otimes {\bSigma_{\cU}^{vv}}^{-1} )\text{vec}(\bmu_{r})\text{vec}(\bmu_{r})^{\T}\big)=\text{vec}({\bmu_r})^{\T}\text{vec}({\bSigma_{\cX}^{-1} \bmu_r \bSigma_{\cU}^{-1}  }) = \text{trace}(\bmu_r^{\T} \bSigma_{\cU}^{-1}\bmu_r \bSigma_{\cX}^{-1} ).$$

\subsection{Proof for Theorem~\ref{prop:prop_1}}
\label{app:prof_for_prop_1}
For any $\bu = (u_1, \dots, u_L)^{\T}$ with $u_1, \dots, u_L \in \cU$, in the prior distribution, 
$\text{vec} \big(\bX_r(\bu, \bv)\big) $ is also normally distributed. We first partition the spatial covariance matrix as 
$$
\begin{bmatrix}
\bSigma_{\cU}^{uu} & \bSigma_{\cU}^{uv} \\ 
\bSigma_{\cU}^{vu} & \bSigma_{\cU}^{vv}
\end{bmatrix},
$$ 
where $\bSigma_{\cU}^{uu}$ and $\bSigma_{\cU}^{vv}$ correspond to the block covariance matrix of $\bu$ and $\bv$, respectively, and $\bSigma_{\cU}^{uv}$ is the cross term. 
Based on this partition, using the formula of conditional multivariate Gaussian distribution, we then have
\begin{align*}
\E_q\Big[\text{vec}\big(\bX_{r}(\bu)\big)\Big]&=
\E_q\Big[\E_q\big[\text{vec}\big(\bX_{r}(\bu)\big) | \bX_{r}(\bv)\big]\Big] \\
&=
\E_q\Big[\E_p\big[\text{vec}\big(\bX_{r}(\bu)\big) | \bX_{r}(\bv)\big]\Big] \\
&= (\bSigma_{\cX} \otimes\bSigma_{\cU}^{uv} )(\bSigma_{\cX} \otimes \bSigma_{\cU}^{vv})^{-1} \E_q \Big[\text{vec}(\bX_r(\bv)\Big] \\&=(\bSigma_{\cX} \otimes \bSigma_{\cU}^{uv})(\bSigma_{\cX}^{-1} \otimes {\bSigma_{\cU}^{vv}}^{-1}) \text{vec}(\bmu_r) \\&= (I \otimes \bSigma_{\cU}^{uv}{\bSigma_{\cU}^{vv}}^{-1} ) \text{vec}(\bmu_r) .
\end{align*}
Therefore,
$\E_q\Big[\text{vec}\big(\bX_{r}(\bu)\big)\Big] = \bSigma_{\cU}^{uv}{\bSigma_{\cU}^{vv}}^{-1}\text{vec}(\bmu_r)$, which means that conditional on $\bSigma_{\cX}$ and $\bSigma_{\cU}$, the mean of variational distribution are mutually independent over factors.

\subsection{Proof for Theorem~\ref{prop:prop_2}}
\label{app:prof_prop_2}
We first derive the variance for the variational distribution of $\bX_r(\bu)$. Note that
\begin{align*}
\var_q\Big[\text{vec}\big(\bX_{r}(\bu)\big)\Big] &= \var_q\Big[\E_q\big[\text{vec}\big(\bX_{r}(\bu)\big) \mid \bX_{r}(\bv)\big]\Big] + \E_q\Big[\var_q\big[\text{vec}\big(\bX_{r}(\bu) \mid \bX_{r}(\bv)\big)\big]\Big].
\end{align*}
The first term is obviously
$$
(I \otimes \bSigma_{\cU}^{uv}{\bSigma_{\cX}^{vv}}^{-1} ) \operatorname{diag}(\bS_{1r}, \dots, \bS_{nr}) .
$$
In an analogy of proof for Theorem~\ref{prop:prop_1}, the second term equals to
\begin{align*}
& 
\bSigma_{\cX} \otimes \bSigma_{\cU}^{uu} -  (\bSigma_{\cX} \otimes \bSigma_{\cU}^{uv}){(\bSigma_{\cX} \otimes \bSigma_{\cU}^{vv} 
)}^{-1}(\bSigma_{\cX} \otimes \bSigma_{\cU}^{uv})^{\T} \\&= \bSigma_{\cX} \otimes \bSigma_{\cU}^{uu} - (\bSigma_{\cX} \otimes \bSigma_{\cU}^{uv})(\bSigma_{\cX}^{-1} \otimes {\bSigma_{\cU}^{vv}}{-1})(\bSigma_{\cX} \otimes \bSigma_{\cU}^{uv})^{\T} \\&=
\bSigma_{\cX} \otimes \bSigma_{\cU}^{uu} - ( I \otimes \bSigma_{\cU}^{uv}{\bSigma_{\cU}^{vv}}^{-1})(\bSigma_{\cX} \otimes \bSigma_{\cU}^{uv})^{\T} \\&=
\bSigma_{\cX} \otimes \bSigma_{\cU}^{uu} - (\bSigma_{\cX} \otimes\bSigma_{\cU}^{uv}{\bSigma_{\cU}^{vv}}^{-1}{\bSigma_{\cU}^{uv}}^{\T}]) \\&=
\bSigma_{\cX}\otimes   (\bSigma_{\cU}^{uu} - \bSigma_{\cU}^{uv}{\bSigma_{\cU}^{vv}}^{-1}{\bSigma_{\cU}^{uv}}^{\T}).
\end{align*}
Therefore, the variance for $\bX_r(u)$ with variational distribution is,
$$
\var_q\Big[\text{vec}\big(\bX_{r}(\bu)\big)\Big] =(I \otimes \bSigma_{\cU}^{uv}{\bSigma_{\cX}^{vv}}^{-1} ) \text{diag}(\bS_{1r}, \dots, \bS_{nr}) + \bSigma_{\cX}\otimes   (\bSigma_{\cU}^{uu} - \bSigma_{\cU}^{uv}{\bSigma_{\cU}^{vv}}^{-1}{\bSigma_{\cU}^{uv}}^{\T}).
$$

\subsection{Proof for Theorem~\ref{prop:prop_3}}
\label{app:prof_prop_3}
Though the model is infinite-dimensional, the inference is conducted on a finite grid of observations. 
Suppose $\{\bY_{t}(\cdot)\}_{1 \leq t \leq n}$ have observations at points $\bu $. Conditional on $\bSigma_{\cX}$ and $\bSigma_{\cU}$, in equation~\eqref{eq:elbo_rewrite} we have
\begin{equation*}
\begin{aligned}
    &\sum_{t=1}^{n} \E_q \Big[ \log p\big(\bY_t(\cdot) \mid \bX_t(\cdot), \bZ, \bA\big) \Big] \\
&=\frac{1}{2\sigma_{\epsilon}^2}\E_q
 \sum_{i=1}^{p} \text{trace} \Big [\big(\bY_i(\bu) - \sum_{r}\beta_{ir} \bX_r(\bu)\big)\big(\bY_i(\bu) -\sum_{r}\beta_{ir} \bX_r(\bu)\big)^{\T}\Big] + \text{constant}\\
&=\frac{1}{2\sigma_{\epsilon}^2}\E_q
  \sum_{i=1}^{p}\sum_{r, j}\text{trace} \Big [\beta_{ir}\beta_{il} \bX_r(\bu)\bX_l(\bu)^{\T}\Big] - 
  \frac{1}{\sigma_{\epsilon}^2}\E_q
  \sum_{i=1}^{p}\sum_{r}\text{trace} \Big [\beta_{ir}\bX_r(\bu)\bY_i(\bu)^{\T}\Big] + \text{constant},
\end{aligned}
\end{equation*}
where $\beta_{ir} = (\bZ \odot \bA)_{ir}$, $\bY_i(\bu)= \begin{bmatrix}
    Y_{1i}(u_1)&   \cdots & Y_{ni}(u_1)   \\
    \cdots& \cdots & \cdots \cdots   \\
    Y_{1i}(u_L) & \cdots & Y_{ni}(u_L)   \\
    \end{bmatrix}, $ and $\bX_r(\bu)= \begin{bmatrix}
    X_{1r}(u_1)&   \cdots & X_{nr}(u_1)   \\
    \cdots& \cdots & \cdots \cdots   \\
    X_{1r}(u_L) & \cdots & X_{nr}(u_L)   \\
    \end{bmatrix}. $
Using the above construction for ${\bX}_r(\cdot)$,  we also have
$$
\E_q \bX_r(\bu)\bX_l(\bu)^{\T} = \left\{\begin{matrix} 
\E_q \tilde{\bX}_r^{(1)}(\bu)\tilde{\bX}_r^{(1)}(\bu)^{\T} + \E_q \tilde{\bX}_r^{(2)}(\bu)\tilde{\bX}_r^{(2)}(\bu)^{\T} \quad r=l,
\\ 
\E_q \tilde{\bX}_r^{(1)}(\bu)\tilde{\bX}_l^{(1)}(\bu)^{\T} \quad \text{otherwise}.
\end{matrix}\right.
$$
and 
$$
\E_q \bX_r(\bu)\bY_i(\bu)^{\T} = \E_q \tilde{\bX}_r^{(1)}(\bu)\bY_i(\bu)^{\T},$$
because $\E_q \tilde{\bX}_r^{(2)}(\bu)=\boldsymbol{0}$,
where $\tilde{\bX}_r^{(1)}(\bu)=\begin{bmatrix}
\tilde{X}_{1r}^{(1)}(u_1)&   \cdots & \tilde{X}_{nr}^{(1)}(u_1)   \\
    \cdots& \cdots & \cdots \cdots   \\
    \tilde{X}_{1r}^{(1)}(u_L) & \cdots & \tilde{X}_{nr}^{(1)}(u_L)   \\
    \end{bmatrix}$ 
    and
$\tilde{\bX}_r^{(2)}(\bu)=\begin{bmatrix}
\tilde{X}_{1r}^{(2)}(u_1)&   \cdots & \tilde{X}_{nr}^{(2)}(u_1)   \\
    \cdots& \cdots & \cdots \cdots   \\
    \tilde{X}_{1r}^{(2)}(u_L) & \cdots & \tilde{X}_{nr}^{(2)}(u_L)   \\
    \end{bmatrix}.$

Furthermore, 
$$
\E_q\text{trace} \big[\tilde{\bX}_r^{(2)}(\bu)\tilde{\bX}_r^{(2)}(\bu)^{\T}\big]  = \text{trace}[\Sigma_{\cX}] \text{trace}[\bSigma_{\cU}^{uu} - \bSigma_{\cU}^{uv}{\bSigma_{\cU}^{vv}}^{-1}{\bSigma_{\cU}^{uv}}^{\T}] .
$$
Given the above results, we obtain that 
\begin{equation*}
\begin{aligned}
    &\sum_{t=1}^{n} \E_q \Big[ \log p\big(\bY_t(\cdot) \mid \bX_t(\cdot), \bZ, \bA\big) \Big] \\
&=\frac{1}{2\sigma_{\epsilon}^2}\E_q
  \sum_{i=1}^{p}\sum_{r, j}\text{trace} \Big [\beta_{ir}\beta_{il} \bX_r^{(1)}(\bu)\bX_l^{(1)}(\bu)^{\T}\Big] - 
  \frac{1}{\sigma_{\epsilon}^2}\E_q
  \sum_{i=1}^{p}\sum_{r}\text{trace} \Big [\beta_{ir}\bX_r^{(1)}(\bu)\bY_i(\bu)^{\T}\Big] \\
  & \quad + \frac{1}{2\sigma_{\epsilon}^2} 
\left \|\bZ \odot \bA\right \|_F^2\text{trace}[\bSigma_{\cX}] \text{trace}[\bSigma_{\cU}^{uu} - \bSigma_{\cU}^{uv}{\bSigma_{\cU}^{vv}}^{-1}{\bSigma_{\cU}^{uv}}^{\T}]  +\text{constant}.
\end{aligned}
\end{equation*}
Therefore, conditional on $\bSigma_{\cX}$ and  $\bSigma_{\cU}$, ELBO is irrelevant to the inter-task component $\bX_r^{(2)}(\bu)$.

\subsection{Derivations for Equation~(\ref{eq:prediction_x})}
\label{app:deri_for_prediction}
$\bar{\bY}_{t+1}(\bu)=(\bar{\bZ} \odot \bar{\bA}) \bar{\bX}_{t+1}(\bu)$ is obvious as the variational variables are assumed to be independent. 

We first compute the predictive mean for the inducing variables at time $n+1$, $\bar{X}_{n+1,r}(\bv)$. In an analogy to Theorem~\ref{prop:prop_1}, as the spatial kernel and temporal kernel are separable, we have
$$
\bar{X}_{n+1,r}(\bv)^{\T} =\bSigma_{\cX}^{n+1, 1:n} \bSigma_{\cX}^{-1} \bmu_{r}^{\T},$$
where $\bSigma_{\cX}^{n+1, 1:n}= \big[
    \kappa_{\cX}(\cX_{n+1}, \cX_{0}), 
    \kappa_{\cX}(\cX_{n+1}, \cX_{1}),     \cdots , \kappa_{\cX}(\cX_{n+1},\cX_{n-1}) , \kappa_{\cX}(\cX_{n+1},\cX_{n})  
    \big]\in \R^{1 \times n}. $
Moreover, we can predict $\bar{X}_{n+1,r}(\bu)$ by
$$
\bar{X}_{n+1,r}(\bu) = \bSigma_{\cU}^{uv}{\bSigma_{\cU}^{vv}}^{-1}\bar{X}_{n+1,r}(\bv) = \bSigma_{\cU}^{uv}{\bSigma_{\cU}^{vv}}^{-1} \bmu_{r}
\bSigma_{\cX}^{-1}  {\bSigma_{\cX}^{n+1, 1:n}}^{\T}$$

\section{Algorithm of Inference }
\label{app:algo}
The steps of Bayesian inference for DF\textsuperscript{2}M are summarized in Algorithm~\ref{alg:alg1} below.

\begin{center}
%\begin{minipage}{0.9\linewidth}
\begin{algorithm}%[H]
\caption{Bayesian Inference for DF\textsuperscript{2}M}
\label{alg:alg1}
%\begin{minipage}{0.7\textwidth}
%\begin{varwidth}[t]{1.0\linewidth}
\begin{algorithmic}
\STATE{%Use FPCA to obtain initial values of factors.\\
Set up initialization of trainable parameters in deep learning models.}
\REPEAT 
    \STATE{
    1. Update variational distribution parameters $\bmu_{tr}$ and $\bS_{tr}$ for inducing variables $\bX_{tr}(\bv)$, along with other variational parameters,
}
\\
    \STATE{
    2. Update trainable parameters in sequential deep learning framework $H$ using the gradient of ELBO with respect to $\bSigma_{\cX}$,}\\

\UNTIL {the convergence of the ELBO  in equation \eqref{eq:elbo_rewrite}}.
\end{algorithmic}
%\end{minipage}
%\end{varwidth}
\end{algorithm}
%\end{minipage}
\end{center}

\section{Dataset and Preprocessing}
\label{app:data}
\textit{Japanese Mortality} dataset is available at \url{https://www.ipss.go.jp/p-toukei/JMD/index-en.html}. We use log transformation and only keep the data with ages less than 96 years old.
\textit{Energy Consumption}  dataset is available at \url{https://data.london.gov.uk/dataset/smartmeter-energy-use-data-in-london-households}. After removing samples with too many missing values, we randomly split the data into 40 groups and take the average to alleviate the impact of randomness. 
\textit{Global Mortality}  dataset downloaded from  \url{http://www.mortality.org/} contains mortality data from 32 countries, we  use log transformation as well and keep the data with age less than 60 years old.
\textit{Stock Intraday} dataset is
obtained from the Wharton Research Data Services (WRDS) database. Link to the codes for the experiments: \url{https://github.com/yiruiliu110/df2m}.
%\newpage
\section{Deep Learning Structures and Hyperparameters}
\label{app:dl_details}

Our deep learning model structure begins with a layer normalization process, designed to standardize the features within each individual sample in a given batch. Following this, the data is fed into a custom linear layer that implements a fully-connected layer alongside a ReLU activation function. The architecture then varies based on the specific model used, with the possibilities including a fully-connected neural network with Relu activation, LSTM, GRU, or an attention mechanism. The final component of the model is a linear layer that translates the output from the LSTM, GRU, or attention mechanism into the final predictions with the desired output size. To ensure an unbiased comparison between DF\textsuperscript{2}M and conventional deep learning models, we configure both to have a \texttt{hidden\_size} of 15 and restrict them to a single layer. For the ATTN model, we also set it to use one head.
We also run experiments using multiple layers and heads with Bayesian hyperparameter optimization and compare the results in Table~\ref{tab:compa_dl}. Compared to  standard deep learning models with multiple layers, DF\textsuperscript{2}M achieves better or comparable results.

\renewcommand{\thetable}{F.\arabic{table}}
\setcounter{table}{0}
\begin{table}[h!]
    \centering
    \caption{The comparison of DF\textsuperscript{2}M to standard deep learning models with multiple layers. For formatting reasons, the standard deviations for MAPEs are multiplied by 10, and the standard deviations for MSPEs are multiplied by 10$^2$, except for \textit{Energy Consumption} dataset. }
    \label{tab:compa_dl}
    \vskip 10pt
     \centering
     \resizebox{0.8\textwidth}{!}{%
  \begin{minipage}{\textwidth}
  \centering
   %\resizebox{0.95\linewidth}{!}{ \setlength{\tabcolsep}{2.5pt}
\begin{tabular}{llccc@{\hspace{10pt}}ccc}
    \toprule
    & \multicolumn{4}{c}{{\scriptsize DF\textsuperscript{2}M}}  &  \multicolumn{3}{c}{Standard Deep learning}  \\
    \cmidrule(l{5pt}r{10pt}){2-5} \cmidrule(l{2pt}r{0pt}){6-8} 
    &     & $h$=1   & $h$=2  &  $h$=3    & $h$=1   & $h$=2  &  $h$=3\\
\midrule\multirow{2}{*}{\rotatebox{0}{\parbox{1.5cm}{\centering{\scriptsize\textbf{\textit{\quad\quad Japanese\\Mortality}}}}}} & MSPE & 3.608 & 3.839 & 3.958 & 3.786 & 4.159 & 4.341  \\
& MAPE & 1.119 & 1.203 & 1.264 & 1.180 & 1.288 & 1.367  \\
%&   method & \multicolumn{3}{c}{{\small DF\textsuperscript{2}M-ATTN}}  &  \multicolumn{3}{c}{{\small LSTM}} 
%\\
\midrule\multirow{2}{*}{\rotatebox{0}{\parbox{1.5cm}{\centering{\scriptsize\textbf{\textit{\quad\quad Energy\\Consm.}}}}}} & MSPE & 8.928 & 11.60 & 17.26 & 9.380 & 11.19 & 12.79  \\
& MAPE & 2.176 & 2.478 & 3.063 & 2.230 & 2.440 & 2.651  \\
\midrule\multirow{2}{*}{\rotatebox{0}{\parbox{1.5cm}{\centering{\scriptsize\textbf{\textit{ Global\\Mortality}}}}}} & MSPE & 7.672 & 8.088 & 8.954 & 8.196 & 8.755 & 9.322  \\
& MAPE & 1.726 & 1.823 & 1.978 & 1.639 & 1.753 & 1.857  \\
\midrule\multirow{2}{*}{\rotatebox{0}{\parbox{1.5cm}{\centering{\scriptsize\textbf{\textit{ Stock\\ Intraday}}}}}} & MSPE & 99.58 & 101.2 & 89.82 & 100.0 & 95.68 & 88.52  \\
& MAPE & 6.424 & 6.505 & 6.269 & 6.450 & 6.283 & 6.162  \\

\bottomrule\end{tabular} 
%}
   \end{minipage}}
\end{table}

We employ Bayesian hyperparameter optimization to tune the key hyperparameters of our model. The tuned  hyperparameters are listed below. The best outcomes for \textit{Japanese Mortality} are reached through a 3-layer LSTM model, which utilizes a dropout rate of 0.07, a learning rate of 0.0008, a weight decay coefficient of 0.0002, and a hidden size of 64. Similarly, for \textit{Energy Consumption}, a 3-layer GRU model providing the best results employs a dropout rate of 0.08, a learning rate of 0.0004, a weight decay coefficient of 0.00009, and a hidden layer size of 64. In the case of \textit{Global Mortality}, the best performance is achieved with a 2-layer GRU model that operates with a dropout rate of 0.33, a learning rate of 0.001, a weight decay coefficient of 0.0002, and a hidden layer size of 48. Lastly, for \textit{Stock Intraday}, the best results are seen with a 5-layer model featuring a 3-head attention mechanism, with a dropout rate of 0.10, a learning rate of 0.0007, a weight decay coefficient of 0.0010, and a hidden layer size of 2.

\newpage
\section{Standard Deviation of the Results}
\label{app:std}
In parallel to the computation of MAPE and MSPE, we calculate their associated standard deviations by
$$
\text{MAPE-STD}(h) \!=\! \Big({\frac{1}{n_2-h}
\sum_{t=n_1+h}^{n}  \big\{\sum_{j=1}^{p} \sum_{k=1}^{K}  \frac{1}{Kp}
 \big| \hat{Y}_{tj}(u_k) - Y_{tj}(u_k) \big| - \text{MAPE}(h)\big\}^2}\Big)^{\frac{1}{2}}, 
 $$
 $$
\text{MSPE-STD}(h) \!=\! \Big({\frac{1}{n_2-h}
\sum_{t=n_1+h}^{n}  \big\{\sum_{j=1}^{p} \sum_{k=1}^{K}  \frac{1}{Kp}
\big[ \hat{Y}_{tj}(u_k) - Y_{tj}(u_k) \big]^2- \text{MSPE}(h)\big\}^2}\Big)^{\frac{1}{2}}.
$$
The findings are presented in Table~\ref{tab:std} below. The results indicate that the DF\textsuperscript{2}M-based methods exhibit a smaller or comparable standard deviation compared to other competitors.

\renewcommand{\thetable}{H.\arabic{table}}
\setcounter{table}{0}
\begin{table}[ht!]
	\centering
	\caption{Standard deviation of  DF\textsuperscript{2}M and Standard Deep Learning Models. For formatting reasons, the standard deviations for MAPEs are multiplied by 10, and the standard deviations for MSPEs are multiplied by 10$^2$, except for \textit{Energy Consumption} dataset. } 
	\label{tab:std}
 \small
 \vskip 10pt
 \resizebox{0.8\textwidth}{!}{%
  \begin{minipage}{\textwidth}
\resizebox{\linewidth}{!}{ \setlength{\tabcolsep}{2.5pt}
\begin{tabular}{lccccccccccccccccc}
    \toprule
    & \multirow{2}{*}{$h$} & \multicolumn{2}{c}{{\scriptsize DF\textsuperscript{2}M-LIN}} & \multicolumn{2}{c}{{\scriptsize LIN}}  & \multicolumn{2}{c}{{\scriptsize DF\textsuperscript{2}M-LSTM}} & \multicolumn{2}{c}{{\scriptsize LSTM}} & \multicolumn{2}{c}{{\scriptsize DF\textsuperscript{2}M-GRU}} &  \multicolumn{2}{c}{{\scriptsize GRU}} & \multicolumn{2}{c}{{\scriptsize DF\textsuperscript{2}M-ATTN}} &  \multicolumn{2}{c}{{\scriptsize ATTN}}  \\
    \cmidrule(lr){3-4} \cmidrule(lr){5-6} \cmidrule(lr){7-8} \cmidrule(lr){9-10} \cmidrule(lr){11-12} \cmidrule(lr){13-14} \cmidrule(lr){15-16} \cmidrule(lr){17-18}
    &     & {\scriptsize MSPE}   & {\scriptsize MAPE}   & {\scriptsize MSPE}   & {\scriptsize MAPE}   & {\scriptsize MSPE}  & {\scriptsize MAPE}   & {\scriptsize MSPE}  & {\scriptsize MAPE}   & {\scriptsize MSPE}  & {\scriptsize MAPE}  & {\scriptsize MSPE}  & {\scriptsize MAPE}   & {\scriptsize MSPE}  & {\scriptsize MAPE}   & {\scriptsize MSPE}  & {\scriptsize MAPE}  \\
   &     & {\scriptsize -STD}   & {\scriptsize -STD}   & {\scriptsize -STD}   & {\scriptsize -STD}   & {\scriptsize -STD}  & {\scriptsize -STD}   & {\scriptsize -STD}  & {\scriptsize -STD}   & {\scriptsize -STD}  & {\scriptsize -STD}  & {\scriptsize -STD}  & {\scriptsize -STD}   & {\scriptsize -STD}  & {\scriptsize -STD}   & {\scriptsize -STD}  & {\scriptsize -STD}  \\
    
\midrule\multirow{3}{*}{\rotatebox{90}{\parbox{1.5cm}{\centering{\tiny\textbf{\textit{\quad Japanese\\\quad Mortality}}}}}} & 1 & 1.794 & 0.179 & 3.909 & 0.757 & 1.687 & 0.168 & 2.180 & 0.197 & 1.988 & 0.198 & 6.578 & 0.608 & 1.780 & 0.178 & 1.017 & 0.107 \\
 & 2 & 1.737 & 0.173 & 4.788 & 0.864 & 1.717 & 0.171 & 2.833 & 0.200 & 2.066 & 0.206 & 5.746 & 0.457 & 1.449 & 0.144 & 1.043 & 0.116 \\
 & 3 & 3.841 & 0.384 & 5.150 & 0.915 & 1.728 & 0.172 & 3.040 & 0.316 & 2.577 & 0.257 & 7.320 & 0.568 & 1.735 & 0.173 & 0.763 & 0.077 \\
\midrule\multirow{3}{*}{\rotatebox{90}{\parbox{1.5cm}{\centering{\tiny\textbf{\textit{\quad  Energy\\\quad  Consm.}}}}}} & 1 & 0.841 & 0.841 & 14.91 & 1.354 & 0.724 & 0.724 & 11.39 & 1.039 & 0.679 & 0.679 & 9.683 & 0.833 & 1.029 & 1.029 & 12.47 & 1.104 \\
 & 2 & 1.134 & 1.134 & 15.32 & 1.297 & 0.846 & 0.846 & 11.98 & 0.979 & 1.121 & 1.121 & 18.41 & 1.407 & 1.203 & 1.203 & 12.67 & 1.082 \\
 & 3 & 1.080 & 1.080 & 16.35 & 1.262 & 1.229 & 1.229 & 12.89 & 1.049 & 0.985 & 0.985 & 13.05 & 0.949 & 1.308 & 1.308 & 12.72 & 1.060 \\
\midrule\multirow{3}{*}{\rotatebox{90}{\parbox{1.5cm}{\centering{\tiny\textbf{\textit{\quad Global\\\quad Mortality}}}}}} & 1 & 3.519 & 0.351 & 14.03 & 1.514 & 0.686 & 0.068 & 3.484 & 0.546 & 1.088 & 0.108 & 4.108 & 0.238 & 1.379 & 0.137 & 1.483 & 0.103 \\
 & 2 & 2.191 & 0.219 & 13.61 & 1.503 & 1.469 & 0.146 & 4.883 & 0.623 & 1.461 & 0.146 & 4.318 & 0.276 & 1.386 & 0.138 & 1.664 & 0.139 \\
 & 3 & 2.580 & 0.258 & 14.03 & 1.466 & 2.676 & 0.267 & 4.803 & 0.640 & 2.365 & 0.236 & 5.747 & 0.269 & 1.386 & 0.138 & 2.602 & 0.217 \\
\midrule\multirow{3}{*}{\rotatebox{90}{\parbox{1.5cm}{\centering{\tiny\textbf{\textit{\quad Stock\\\quad Intraday}}}}}} & 1 & 20.73 & 2.073 & 99.88 & 2.720 & 21.75 & 2.175 & 117.6 & 2.858 & 18.87 & 1.887 & 291.3 & 4.987 & 18.93 & 1.893 & 77.01 & 2.058 \\
 & 2 & 22.27 & 2.227 & 86.99 & 2.361 & 27.17 & 2.717 & 93.25 & 1.823 & 19.63 & 1.963 & 329.2 & 4.917 & 21.25 & 2.125 & 78.82 & 2.124 \\
 & 3 & 18.80 & 1.880 & 109.2 & 2.989 & 26.59 & 2.659 & 115.7 & 2.614 & 18.85 & 1.885 & 305.7 & 5.071 & 20.78 & 2.078 & 79.17 & 2.184 \\
\bottomrule\end{tabular} }
\end{minipage}}

\end{table}

%\begin{enumerate}
%    \item \texttt{hidden\_size}: The number of neurons in the hidden layer, with an integer search space ranging from 10 to 100.
%    \item \texttt{dropout}: The dropout rate applied to the hidden layer, with a continuous search space ranging from 0.1 to 0.5.
%    \item \texttt{lr}: The learning rate for the optimization algorithm, with a log-uniform search space spanning from 1e-4 to 1e-2.
%    \item \texttt{weight\_decay}: The weight decay coefficient used in the optimization algorithm, with a log-uniform search space ranging from 1e-5 to 1e-3. 
%\end{enumerate}

%We leverage the Optuna library to perform Bayesian HPO, which intelligently navigates the search space by learning from the performance of previous hyperparameter configurations. 

%\renewcommand{\thetable}{F.\arabic{table}}
%\setcounter{table}{0}
%\begin{table}[]
%    \centering
%    \caption{Hyperparameters for Deep Learning Models}
%    \label{tab:my_label}
%    \input{tables/hpo_para}
%\end{table}

%\renewcommand{\thetable}{F.\arabic{table}}
%\setcounter{table}{0}
%\begin{table}[]
%    \centering
%    \caption{Hyperparameters for Deep Learning Models}
%    \label{tab:tuned_dl}
%    \input{tables/tuned_dl}
%\end{table}
\end{document}